\documentclass[10pt,twocolumn,letterpaper]{article}

\usepackage[pagenumbers]{cvpr} 

\usepackage{multirow} 
\usepackage{makecell} 

\definecolor{cvprblue}{rgb}{0.21,0.49,0.74}
\usepackage[pagebackref,breaklinks,colorlinks,citecolor=cvprblue]{hyperref}
\usepackage{tabularx}
\usepackage{array} 
\newcolumntype{C}{>{\centering\arraybackslash}X}
\usepackage{amssymb}
\usepackage{ragged2e}
\usepackage{verbatim}

\usepackage{lineno}

\def\paperID{*****} 
\def\confName{CVPR}
\def\confYear{2025}

\title{Controllable Reference-Guided Diffusion with Local–Global Fusion for Real-World
Remote Sensing Image Super-Resolution}

\author{Ce Wang, Wanjie Sun\thanks{Corresponding author} \\
School of Remote Sensing and Information Engineering, Wuhan University\\ Wuhan 430079, China\\
\tt\small {\{cewang, sunwanjie\}@whu.edu.cn}
}
\begin{document}
\maketitle
\begin{abstract}
Super-resolution (SR) techniques can enhance the spatial resolution of remote sensing images, enabling more efficient large-scale earth observation applications. While single-image super-resolution methods enhance low-resolution (LR) images, they neglect valuable complementary information from auxiliary data. Reference-based super-resolution (RefSR) can be interpreted as an information fusion task, where historical high-resolution (HR) reference images are combined with current LR observations. However, existing RefSR methods struggle with real-world complexities, such as cross-sensor resolution gap and significant land cover changes, often leading to under-generation or over-reliance on reference image. To address these challenges, we formulate the RefSR task as a multi-sensor fusion problem and propose CRefDiff, a novel controllable reference-guided diffusion model for real-world remote sensing image super-resolution. To address the under-generation problem, CRefDiff leverages a powerful generative prior to produce accurate structures and textures. To mitigate over-reliance on the reference, we introduce a dual-branch fusion mechanism that adaptively fuse both local and global information from the reference image. Moreover, this novel dual-branch design enables reference strength control during inference, enhancing the model’s interactivity and flexibility. Finally, a strategy named Better Start is proposed to significantly reduce the number of denoising steps, thereby accelerating the inference process. To support further research, we introduce Real-RefRSSRD, a new real-world RefSR dataset for remote sensing images, consisting of HR NAIP and LR Sentinel-2 image pairs with diverse land cover changes and significant temporal gaps. Extensive experiments on Real-RefRSSRD show that CRefDiff achieves state-of-the-art performance across various metrics and improves downstream tasks such as scene classification and semantic segmentation. The dataset and code is publicly available at https://github.com/wwangcece/CRefDiff
\end{abstract}
\section{Introduction}
In the domain of earth observation, high-resolution (HR) and low-resolution (LR) satellite imagery serve complementary roles across a wide range of applications. HR images, with their fine spatial detail, are typically utilized in tasks such as building extraction \citep{ji2018fully}, urban planning \citep{mathieu2007mapping}, and small object detection \citep{rabbi2020small}. In contrast, LR satellite images benefit from lower acquisition costs and higher temporal resolution, making them well suited for applications including change detection \citep{wang2022graph}, agricultural monitoring \citep{segarra2020remote}, and natural disaster assessment \citep{varghese2021reviewing}. However, due to hardware limitations, the advantages of HR and LR images are often mutually exclusive. This has led to a strong demand for software-based algorithms to enhance the spatial resolution of LR images.

Super-resolution (SR) technology aims to reconstruct HR images from LR observations, thereby enabling the acquisition of HR imagery beyond the inherent spatial sampling limitations of digital imaging systems. By providing an efficient and cost-effective approach to exploiting globally accessible LR imagery \citep{jiang2019edge}, SR has become a major research focus in both the remote sensing and computer vision communities \citep{dong2022real, qiu2023cross, xiao2024frequency}. In the early stages of SR development, a variety of approaches were proposed, including reconstruction-based \citep{sun2008image}, example-based \citep{freeman2002example}, sparse representation-based \citep{yang2010image}, and regression-based methods \citep{timofte2013anchored}. With the advent of deep learning, numerous network architectures have been introduced to enhance image reconstruction performance, such as convolutional neural networks (CNNs) \citep{dong2015image}, residual networks \citep{kim2016accurate}, dense networks \citep{zhang2018residual}, and transformer-based architectures \citep{liang2021swinir}. However, studies have shown that relying solely on regression losses such as mean squared error (MSE) often results in over-smoothed reconstructions \citep{wang2020deep}. This limitation stems from the inherently ill-posed nature of the SR task, where simple pixel-wise supervision tends to drive models toward predicting the average of all plausible solutions. To address this issue, various generative models capable of learning the conditional probability distribution of HR images given LR inputs have been introduced into the SR domain. These include Generative Adversarial Networks (GANs) \citep{wang2018esrgan, jiang2019edge}, Normalizing Flows (NFs) \citep{lugmayr2020srflow}, and Diffusion Models (DMs) \citep{saharia2022image}, all of which have significantly improved the perceptual quality of the reconstructed images.

However, when dealing with large upscaling factors, the aforementioned generative approaches often suffer from uncontrollable hallucination issues \citep{zhang2020texture}, where the model produces unrealistic or non-existent textures during the SR process. To mitigate this issue, recent studies have begun to shift from single-image SR (SISR) to reference-based SR (RefSR). By incorporating additional HR reference images, RefSR enables the reconstruction of high-frequency details with improved fidelity. For natural images, the reference is usually obtained from different viewpoints of the same object \citep{zhang2019image, yang2020learning} or is retrieved using web search engines \citep{jiang2021robust}. In the context of remote sensing imagery, Dong et al. \citep{dong2021rrsgan} proposed using HR images from platforms such as Google Earth as references and constructed the first RefSR dataset for remote sensing imagery. However, these datasets generate LR images using bicubic downsampling from corresponding HR images, without accounting for the complex degradations introduced by cross-sensor differences in real-world scenarios. In contrast, we formulate the RefSR task as a fusion problem of heterogeneous remote sensing imagery, where the historical reference images, current LR observations, and current HR targets are all derived from real remote sensing observation:
\begin{equation}
    \begin{aligned}
    I_{\mathrm{hr}}=f_{\theta}\left(I_{\mathrm{lr}},I_{\mathrm{ref}}\right)
    \end{aligned}
    \label{equ:refsr}
\end{equation}
Here, $I_{\mathrm{ref}}$ and $I_{\mathrm{hr}}$ denote the historical and current observations from a high-resolution satellite, while $I_{\mathrm{lr}}$ represents the current observation from a low-resolution satellite. Therefore, once trained, the model $f_{\theta}$ can generate the current HR observation by integrating a historical HR image with frequently acquired LR observations, thereby enabling high spatiotemporal resolution Earth observation.

\begin{figure}[htb]
    \centering
    \includegraphics[width=0.95\linewidth]{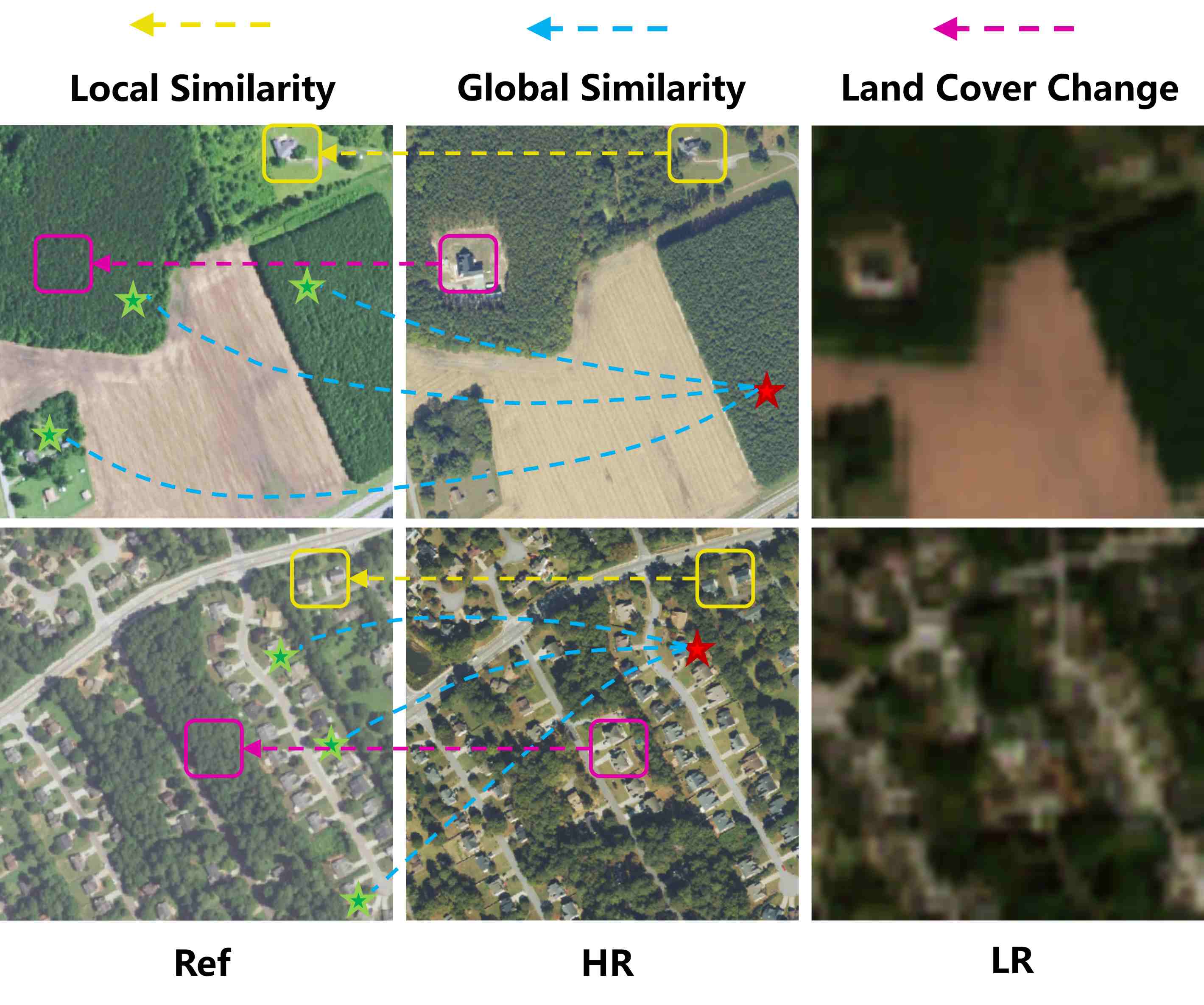}
    \caption{An example data pair from the Real-RefRSSRD. Among the $I_{\mathrm{ref}}$ and $I_{\mathrm{hr}}$ images, three types of relationships can be observed: local similarity in unchanged regions, global similarity, and land cover changes. These priors can provide valuable guidance for the design of SR models.}
    \label{fig:refsr}
\end{figure}

To advance the research on reference-based real-world remote sensing image super-resolution formulated in Eq. (\ref{equ:refsr}), we construct a dataset named Real-RefRSSRD. Specifically, we collect HR imagery from NAIP (with a ground sampling distance of 1 m) and LR imagery from Sentinel-2 (with a ground sampling distance of 10 m). By filtering the images based on their acquisition dates, we ensure a significant temporal interval between the acquisition of $\left(I_{\mathrm{hr}}, I_{\mathrm{lr}}\right)$ and $I_{\mathrm{ref}}$, resulting in notable land cover changes that increase the task's complexity and reflect real-world application scenarios. As shown in Fig. \ref{fig:refsr}, we present a sample image pair from the Real-RefRSSRD. It can be observed that there are certain changes in land cover between $I_{\mathrm{ref}}$ and $I_{\mathrm{hr}}$. In the unchanged regions, there exists strong local similarity between the two images. In the changed regions, despite the differences, the objects in the $I_{\mathrm{hr}}$ still exhibit a global layout similarity to those in the $I_{\mathrm{ref}}$. This global contextual information can also provide valuable cues for SR reconstruction. Therefore, the key challenges in real-world RefSR for remote sensing imagery lie in the following two aspects: 
\begin{enumerate}
    \item Accurate prediction of changes in land cover. The model must effectively identify changed regions by leveraging both the LR observation and the reference image, in order to avoid being misled by outdated or inconsistent information in the reference, which may cause hallucinated content.
    \item Effective utilization of both local and global similarities. The model should fully utilize both local fine-grained details and global contextual information provided by the reference image, thereby compensating for the information loss inherent in the LR observation.
\end{enumerate}
However, most existing approaches rely on end-to-end training pipelines without explicitly modeling the spatial distribution of reference information utilization, making it difficult to handle land cover changes \citep{dong2021rrsgan}. Moreover, they rely solely on either local similarity \citep{tu2024rgtgan} or global similarity \citep{yang2020learning} between the reference and HR images, which limits their ability to effectively address the aforementioned challenges.

To address the aforementioned challenges, we propose a Controllable Reference-Guided Diffusion (CRefDiff) model tailored for the real-world RefSR task. Built upon the pretrained Stable Diffusion (SD) model \citep{rombach2022high}, CRefDiff leverages its strong generative capacity to bridge the substantial resolution gap (exceeding $8\times$) between different satellite platforms. To more effectively fuse both local and global information from the reference image, we introduce a Change Aware Attention (CAA) block and a Semantic Token Aggregation (STA) module, enabling the model to incorporate reference cues at both local and global levels. To alleviate the heavy computational burden caused by the iterative sampling process of diffusion models, we design a Better Start strategy that modifies the starting point of the reverse process, effectively shortening the denoising trajectory. To fundamentally mitigate the model's over-reliance on the reference image, we further introduce a reference strength control mechanism during inference. This mechanism allows users to adaptively adjust the degree to which the model fuses reference information, thereby enhancing both the interactivity and flexibility of the model. The main contributions of this work are summarized as follows:
\begin{enumerate}
    \item We construct Real-RefRSSRD, a real-world reference-based super-resolution dataset for remote sensing. It contains paired real-world HR (NAIP) and LR (Sentinel-2) satellite images across diverse geographic regions, temporal spans (6–14 years), and land cover types. This dataset reflects the real-world challenges of cross-sensor variation and significant land cover change, providing a rigorous benchmark for RefSR models.
    \item We propose CRefDiff, a novel controllable RefSR framework built on pretrained generative diffusion models. It effectively fuses multi-sensor heterogeneous information and mitigates hallucinations through a carefully designed fusion architecture that jointly models local textures and global contextual cues.
    \item We design two novel inference-time mechanisms, Better Start and Reference Strength Control. Better Start shortens the diffusion trajectory, significantly accelerating inference. While, Reference Strength Control allows users to dynamically modulate the influence of the reference image, enhancing fusion flexibility and robustness under different changing conditions.
    \item Extensive experiments on Real-RefRSSRD show that CRefDiff achieves state-of-the-art performance both quantitatively and qualitatively. Moreover, it also improves performance in downstream tasks such as scene classification and semantic segmentation, highlighting its practical value for remote sensing applications.
\end{enumerate}
\section{Related work}
\subsection{Remote Sensing Image Super-Resolution}
In recent years, with the rapid advancement of deep learning techniques, numerous deep learning-based SR methods have been proposed, significantly outperforming traditional approaches \citep{sun2008image, yang2010image, timofte2013anchored, freeman2002example}. Dong et al. \citep{dong2015image} were the first to introduce neural networks into the SR task. Since then, extensive efforts have been devoted to improving SR performance by designing more effective network architectures, such as residual blocks \citep{ledig2017photo}, dense blocks \citep{kim2016accurate}, and recursive blocks \citep{kim2016deeply}. Compared with natural images, remote sensing imagery possesses unique characteristic \citep{liu2023efficient, zou2025hyperspectral}. Consequently, numerous studies have adapted network architectures to better suit these properties, such as multi-scale networks \citep{lei2021hybrid, chen2023msdformer, lei2017super}, multi-stage architectures \citep{lei2021transformer, li2022dual}, gradient guidance mechanisms \citep{jiang2019edge, qiu2023cross}, scene-adaptive networks \citep{li2024local, zhang2020scene} and degradation-adaptive networks \citep{xiao2023degrade}.

Transformers, known for capturing long-range dependencies, have shown strong potential in SR tasks \citep{liang2021swinir, chen2023activating}. Kang et al. \citep{kang2024efficient} proposed a group-wise multi-window self-attention block that reduces computation while maintaining high-quality reconstruction. To address the redundancy of irrelevant tokens and the lack of multi-scale representation, Xiao et al. \citep{xiao2024ttst} introduced the top-k token selective transformer. For hyperspectral image SR, Zhang et al. \citep{zhang2023essaformer} replaced standard attention with a spectral correlation-guided mechanism to embed spectral priors and enhance feature representation.

Despite architectural advances, pixel-wise losses like MSE tend to produce overly smooth outputs by regressing toward the mean of the solution space. To restore fine-grained details, generative models have been widely adopted. Early methods employed GANs to improve perceptual quality \citep{ledig2017photo, wang2018esrgan}, with subsequent variants incorporating multi-attention \citep{jia2022multiattention}, multi-scale attention \citep{wang2023msagan}, and saliency discrimination \citep{ma2019sd}. However, GANs suffer from training instability and mode collapse, prompting interest in alternative generative approaches such as normalizing flows \citep{lugmayr2020srflow, liang2021hierarchical, wu2023conditional} and autoregressive models \citep{guo2022lar, wang2019multi}.

Recently, denoising diffusion probabilistic models \citep{ho2020denoising} (DDPMs) have demonstrated remarkable advantages in the field of AI-generated content (AIGC). Compared to GANs or normalizing flow models, DDPMs offer more stable training, greater architectural flexibility, and better scalability \citep{yang2023diffusion}. As a result, an increasing number of researchers have begun to explore the application of DDPMs to the SR task. Saharia et al. \citep{saharia2022image} were the first to employ the DDPM architecture for image SR, achieving performance comparable to that of GAN-based methods. Yue et al. \citep{yue2024resshift} introduced the concept of residual diffusion to better exploit the available pixels in LR images, significantly reducing inference time. Xiao et al. \citep{xiao2023ediffsr} were the first to apply diffusion models to remote sensing image SR and proposed an efficient denoising network tailored for this task. To further enhance the robustness and generalization of SR models in unseen scenarios, an increasing number of studies have begun to develop SR methods based on large-scale text-to-image pretrained models, such as Stable Diffusion \citep{rombach2022high}. Wang et al. \citep{wang2023exploiting} were the first to explore this direction, significantly improving the perceptual quality of reconstruction results. Khanna et al. \citep{khanna2023diffusionsat} fine-tuned a pretrained SD model using remote sensing imagery and associated metadata, achieving $10\times$ SR for Sentinel-2 images.

Although generative models alleviate over-smoothing issues, they can also introduce hallucinated content, generating unrealistic textures or structures. Unlike previous approaches, our method incorporates reference images and fully exploits both their local fine-grained details and global contextual information. Moreover, it allows users to explicitly control the proportion of information derived from the LR and reference images, thereby enabling more flexible and controllable content generation.

\subsection{Reference-Based Super-Resolution}
Unlike SISR, reference-based super-resolution (RefSR) enhances reconstruction by utilizing both the LR input and an additional HR reference image. In natural image scenarios, references are typically sourced from alternative viewpoints \citep{zheng2018crossnet}, adjacent video frames \citep{caballero2017real}, or retrieved via search engines \citep{jiang2021robust}. These references often exhibit substantial viewpoint disparities from the ground truth, making effective alignment the core challenge of RefSR. Existing alignment strategies can be broadly categorized into explicit alignment via patch matching \citep{zhang2019image, yang2020learning, xie2020feature, lu2021masa} and implicit alignment via deformable convolutions \citep{shim2020robust, jiang2021robust, zhang2022rrsr, cao2022reference}. Patch matching treats LR patches as queries and reference patches as keys/values, aggregating high-frequency details based on patch-wise similarity. For example, Zhang et al. \citep{zhang2019image} performed multi-scale patch matching in a pretrained semantic space, while Lu et al. \citep{lu2021masa} introduced a coarse-to-fine strategy to reduce computational cost. Jiang et al. \citep{jiang2021robust} further improved matching accuracy via contrastive learning and knowledge distillation. To tackle the shortcomings of explicit patch matching, Sun et al. \citep{sun2022learning} proposed a test-time RefSR method, which learns a reference image specific codebook as prior and super-resolve the input LR image by searching the nearest code to replace the LR image feature. In deformable convolution-based methods, initial sampling offsets are typically guided by patch matching and refined adaptively based on content. Their flexibility and content awareness generally lead to superior alignment accuracy \citep{chan2021understanding}. Shim et al. \citep{shim2020robust} first replaced patch matching with deformable convolutions, and Cao et al. \citep{cao2022reference} introduced a top-k deformable attention mechanism to suppress irrelevant tokens. Huang et al. \citep{huang2022task} integrated both strategies for more precise alignment. Beyond external references, some approaches explore self-reference. Shocher et al. \citep{shocher2018zero} leveraged internal recurrence within a single image, while Li et al. \citep{li2023self} used a pretrained Stable Diffusion model to generate a high-frequency reference for SR.

Unlike natural images, satellite imagery is typically acquired in a nadir-looking manner. Dong et al. \citep{dong2021rrsgan} proposed using publicly available historical HR imagery from Google Earth as reference, highlighting that the key challenge in remote sensing RefSR is not viewpoint variation but potential land cover changes. To address the domain gap caused by varying imaging conditions, Min et al. \citep{min2023bridging} employed style transfer techniques to reduce distributional discrepancies between reference and target images, thereby improving texture transfer fidelity. Zhang et al. \citep{zhang2023reference} proposed a feature compression module to extract reference features, thereby reducing the dependency on network depth. Wang et al. \citep{wang2024reference} designed a coarse-to-fine matching strategy and introduced a threshold-based filtering mechanism to mitigate the impact of mismatches. Guo et al. \citep{guo2024dtesr} proposed a dynamic texture guidance mechanism to enhance the efficiency of reference information utilization. The work most relevant to ours is \citep{he2021spatial}, which introduced a pixel-wise generative model conditioned on spatiotemporal coordinates to fuse current LR observations with historical HR imagery. However, this method only exploits local information from reference images and lacks explicit control over reference influence during inference. Beyond historical images, cross-modal references such as hyperspectral imagery \citep{meng2022large}, vector maps \citep{wang2025semantic}, and semantic masks \citep{dong2024building} have also been explored to boost SR performance. In this work, we propose a controllable generative framework that jointly leverages the intrinsic priors of pretrained diffusion models and both local and global cues from reference images to generate high-fidelity spatiotemporal SR results.
\section{Preliminary}   
\subsection{Problem Statement}
\label{sec:naive_solution}
\begin{figure}[htb]
    \centering
    \includegraphics[width=\linewidth]{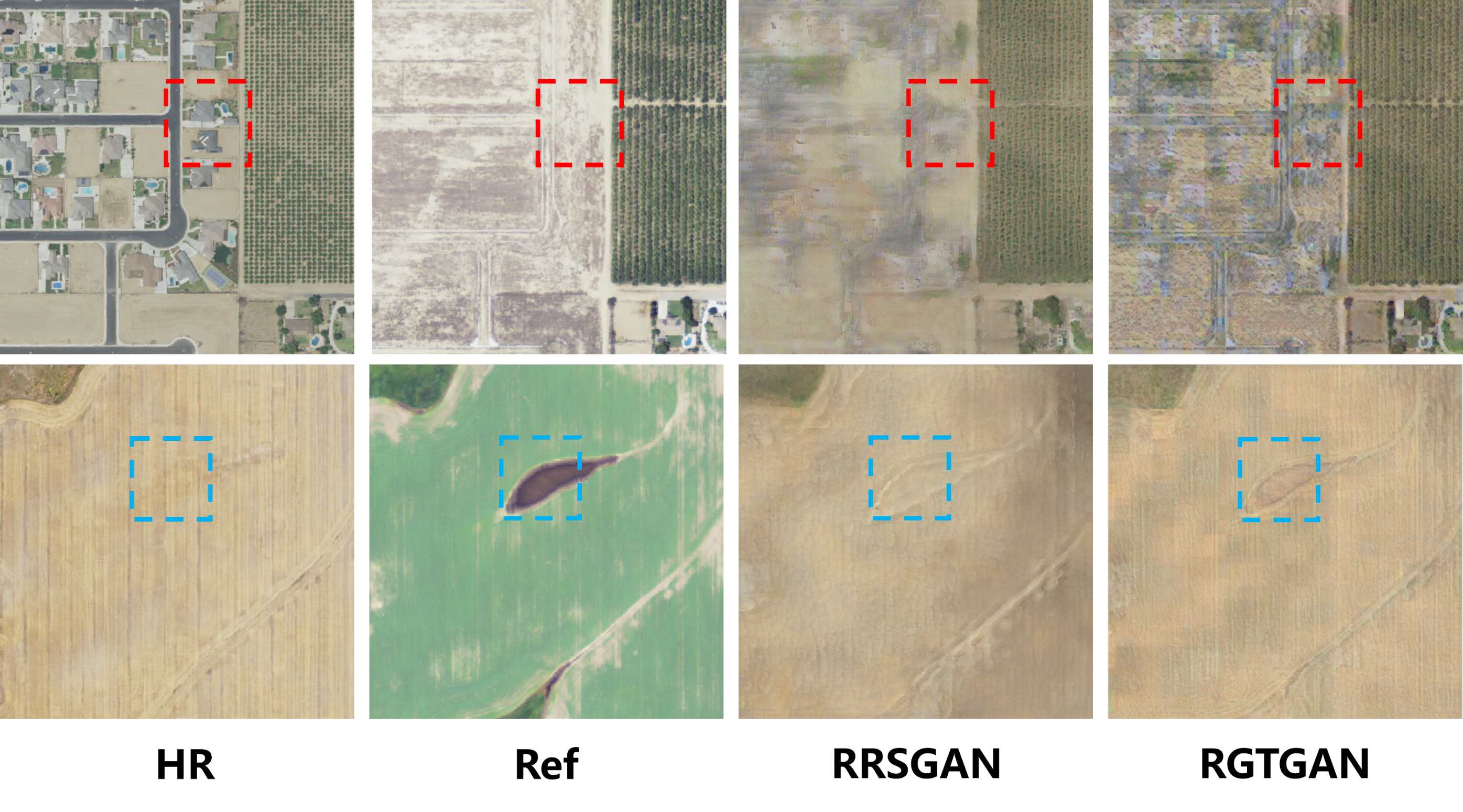}
    \caption{In the presence of substantial land cover changes, current RefSR methods often exhibit issues of under-generation (failing to recover new structures) and over-reliance on the reference (hallucinating outdated or incorrect content).}
    \label{fig:gene_vs_ref}
\end{figure}

\begin{figure}[htb]
    \centering
    \includegraphics[width=\linewidth]{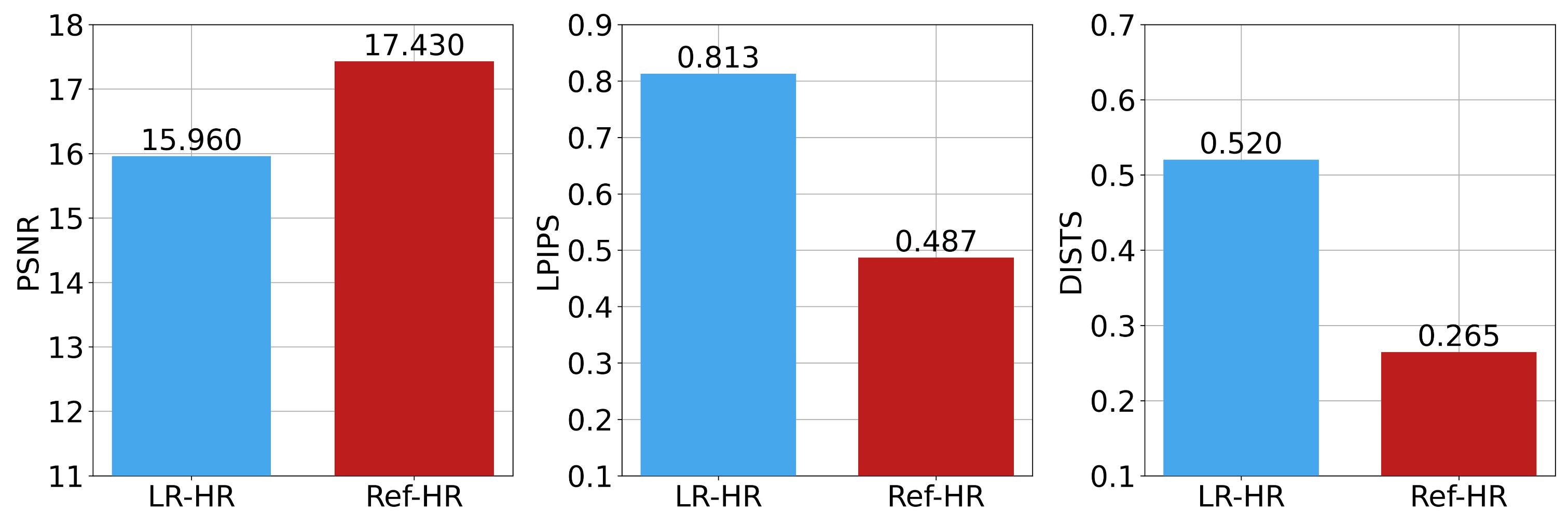}
    \caption{We calculate three similarity metrics between LR-HR and Ref-HR image pairs: PSNR, LPIPS, and DISTS. For PSNR, the higher, the better. While for LPIPS and DISTS, the lower, the better. In the figure, red indicates the best results.}
    \label{fig:lr-ref-compare}
\end{figure}

Unlike natural images, reference satellite images are usually coarsely aligned with HR images in the spatial domain. Therefore, to fuse the information from a roughly aligned reference image and a LR image, we train and test existing RefSR methods for remote sensing imagery (in our case, RRSGAN \citep{dong2021rrsgan}) and RGTGAN \citep{tu2024rgtgan}) on the collected remote sensing imagery. However, we observe these approaches struggle to handle regions with significant land cover changes. As illustrated in Fig. \ref{fig:gene_vs_ref}, when new objects appear in the target HR image, the model, relying primarily on the LR input, fails to reconstruct accurate structures, a phenomenon we refer to as \textbf{under-generation}. Conversely, when certain objects present in the reference image are missing in the HR image, the model tends to over-rely on the reference and hallucinates incorrect content, which we term the \textbf{over-reliance} problem.  As shown in Fig. \ref{fig:lr-ref-compare}, we present three similarity metrics between the ground truth (GT) image and the reference and LR images. Despite the presence of land cover changes, the reference image consistently exhibits a higher similarity to the GT image across all metrics. This demonstrates that the changed regions between the reference and HR images are notably fewer than the unchanged ones, which leads the network to favor a direct ``copy-and-paste'' strategy from the reference image \citep{yang2023paint}. This suggests that existing methods fail to capture the underlying discrepancies between LR and reference images, thereby limiting their ability to handle regions with substantial land-cover changes during inference.
To address the aforementioned issues, we propose three key strategies: 1) We leverage the powerful generative capability of a pretrained diffusion model to mitigate the under-generation problem. 2) We apply aggressive data augmentation to the reference images to alleviate the issue of over-reliance. 3) We enable user-controllable adaptation of the model's dependence on the reference during inference, thereby enhancing the interactivity and flexibility of the system. 

\subsection{Diffusion Model}
Diffusion models (DMs)~\citep{ho2020denoising} are a class of generative models that learn data distributions by simulating a Markovian forward-noising process and a learned reverse denoising process.

\subsubsection{Forward Process}
In the forward process, a clean sample $x_0$ is gradually corrupted by Gaussian noise over $T$ steps, forming a Markov chain:
\begin{equation}
q(x_t | x_{t-1}) = \mathcal{N}(x_t; \sqrt{1 - \beta_t} \, x_{t-1}, \beta_t I),
\end{equation}
where $\{\beta_t\}_{t=1}^T$ is a predefined variance schedule. The marginal distribution can be written as:
\begin{equation}
q(x_t | x_0) = \mathcal{N}(x_t; \sqrt{\bar{\alpha}_t} \, x_0, (1 - \bar{\alpha}_t) I),
\end{equation}
with $\bar{\alpha}_t = \prod_{s=1}^t (1 - \beta_s)$.

\subsubsection{Reverse Process}
The generative process aims to recover $x_0$ from Gaussian noise $x_T \sim \mathcal{N}(0, I)$ by learning the reverse transitions:
\begin{equation}
p_\theta(x_{t-1} | x_t) = \mathcal{N}(x_{t-1}; \mu_\theta(x_t, t), \Sigma_\theta(x_t, t)).
\end{equation}
In practice, the model often predicts the added noise $\epsilon$ instead of directly modeling $\mu_\theta$ \citep{ho2020denoising}, leading to a simplified training objective:
\begin{equation}
\mathcal{L}_{\text{diff}} = \mathbb{E}_{x_0, t, \epsilon}  \| \epsilon - \epsilon_\theta(x_t, t) \|^2.
\end{equation}

\subsubsection{Stable Diffusion}
Stable Diffusion (SD)~\citep{rombach2022high} extends diffusion models to high-resolution image synthesis by operating in a compressed latent space. It first employs a pretrained Variational Autoencoder (VAE) to encode the image into a lower-dimensional latent representation $z$, on which the diffusion process is performed:
\begin{equation}
z_0 = \mathcal{E}(x_0), \quad z_T \sim \mathcal{N}(0, I), \quad \hat{z}_0 \leftarrow \text{DDPM}(z_T),
\end{equation}
Here, DDPM refers to the reverse sampling process. The denoised latent $z_0$ is then decoded back into the image space:
\begin{equation}
\hat{x}_0 = \mathcal{D}(\hat{z}_0).
\end{equation}
By shifting diffusion from pixel space to the latent space of the VAE, SD significantly improves sampling efficiency and enables high-quality generation.

\begin{figure*}[htb]
    \centering
    \includegraphics[width=\linewidth]{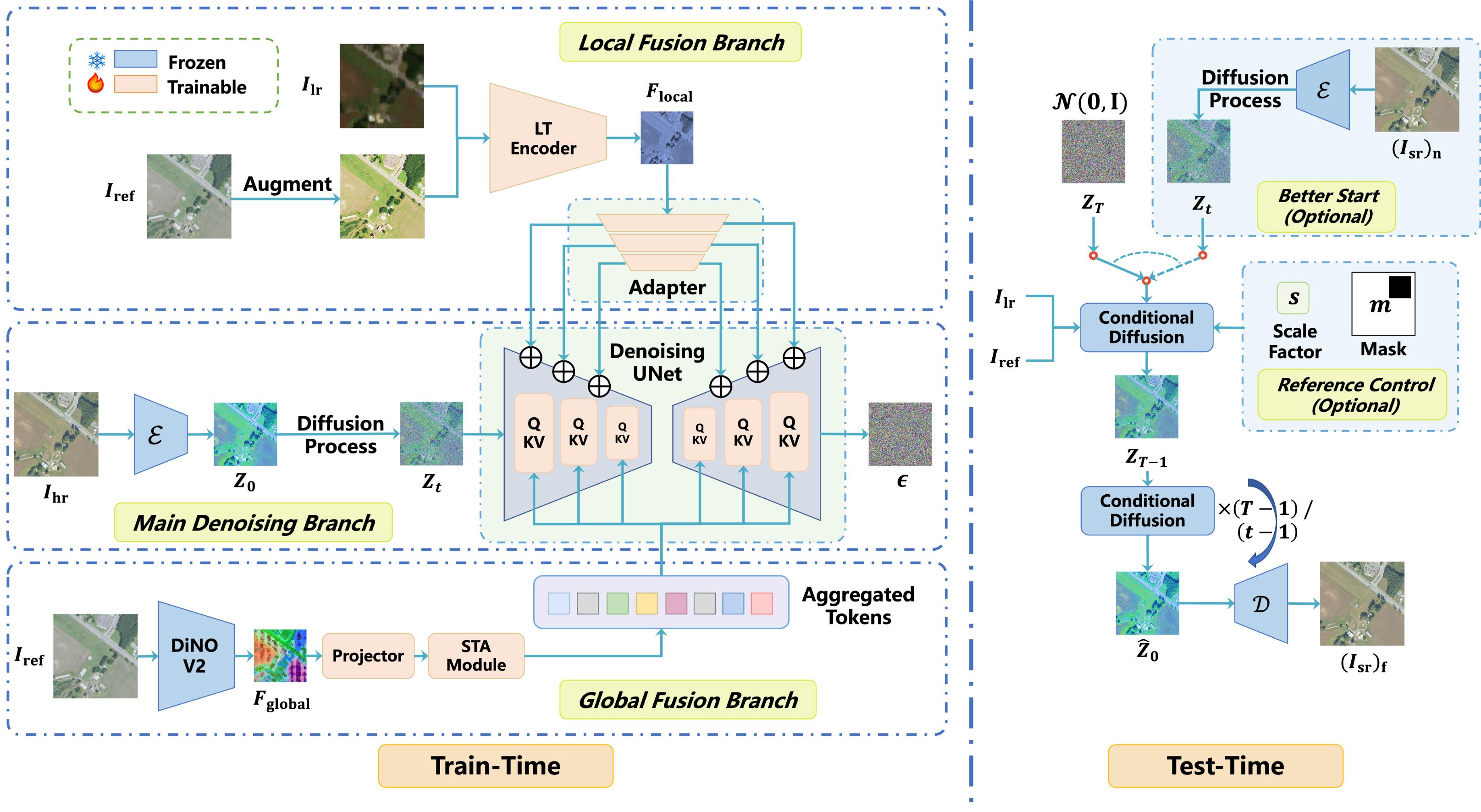}
    \caption{The framework of CRefDiff consists of two fusion branches to guide the UNet in denoising. The local fusion branch extracts texture-rich multi-scale features from $I_\mathrm{lr}$ and $I_\mathrm{ref}$ using a local texture encoder and an adapter. The global fusion branch employs DiNOv2 and a semantic token aggregator to produce semantically rich tokens from $I_\mathrm{ref}$. During inference, the better start strategy reduces reverse steps, while reference strength control adjusts the model's reliance on reference information.}
    \label{fig:arch}
\end{figure*}
\section{Method}
\subsection{Overview}
Fig. \ref{fig:arch} illustrates the overall architecture of our proposed CRefDiff framework, which consists of three branches: a local fusion branch, a global fusion branch, and the main denoising branch. In the local fusion branch, the LR image and the reference image are fused via a Local Texture Encoder (LT-Encoder) to extract local features $F_{\mathrm{local}}$. Local features are then passed through a multi-scale adapter module to generate hierarchical guidance for the UNet during the denoising process. In the global fusion branch, global features $F_{\mathrm{global}}$ are extracted from the reference image using the DiNOv2 model \citep{oquab2023dinov2}. A Semantic Token Aggregation (STA) module is then employed to distill semantic tokens from these features. These tokens are injected into the UNet through cross-attention, providing global semantic guidance throughout the denoising process.

During inference, we introduce two optional strategies: Better Start and Reference Strength Control. The Better Start strategy leverages the output of a lightweight fusion model as the initial state for the reverse diffusion process. This significantly shortens the denoising trajectory and reduces the number of required inference steps. The Reference Strength Control strategy enables users to specify either a scalar $s$ or a spatial mask $m$ to control the degree to which reference information influences the model, at both global and local levels. This design fundamentally mitigates the over-reliance problem by allowing adaptive regulation of reference usage.

\subsection{Local Fusion Branch}
The local fusion branch is designed to fully exploit the local similarity between the reference image and the underlying GT image. It takes both $I_{\mathrm{lr}}$ and $I_{\mathrm{ref}}$ as inputs and outputs a set of multi-scale features to guide the denoising process of the UNet. Specifically, this branch consists of two modules: a local texture encoder and an adapter. The encoder aims to effectively fuse information from $I_{\mathrm{lr}}$ and $I_{\mathrm{ref}}$, producing a tensor $F_\mathrm{local}$. The adapter then transforms $F_\mathrm{local}$ into a set of multi-scale features that are compatible with the UNet architecture, aligning them with the intrinsic priors of the pre-trained Stable Diffusion model.

\begin{figure}[htb]
    \centering
    \includegraphics[width=\linewidth]{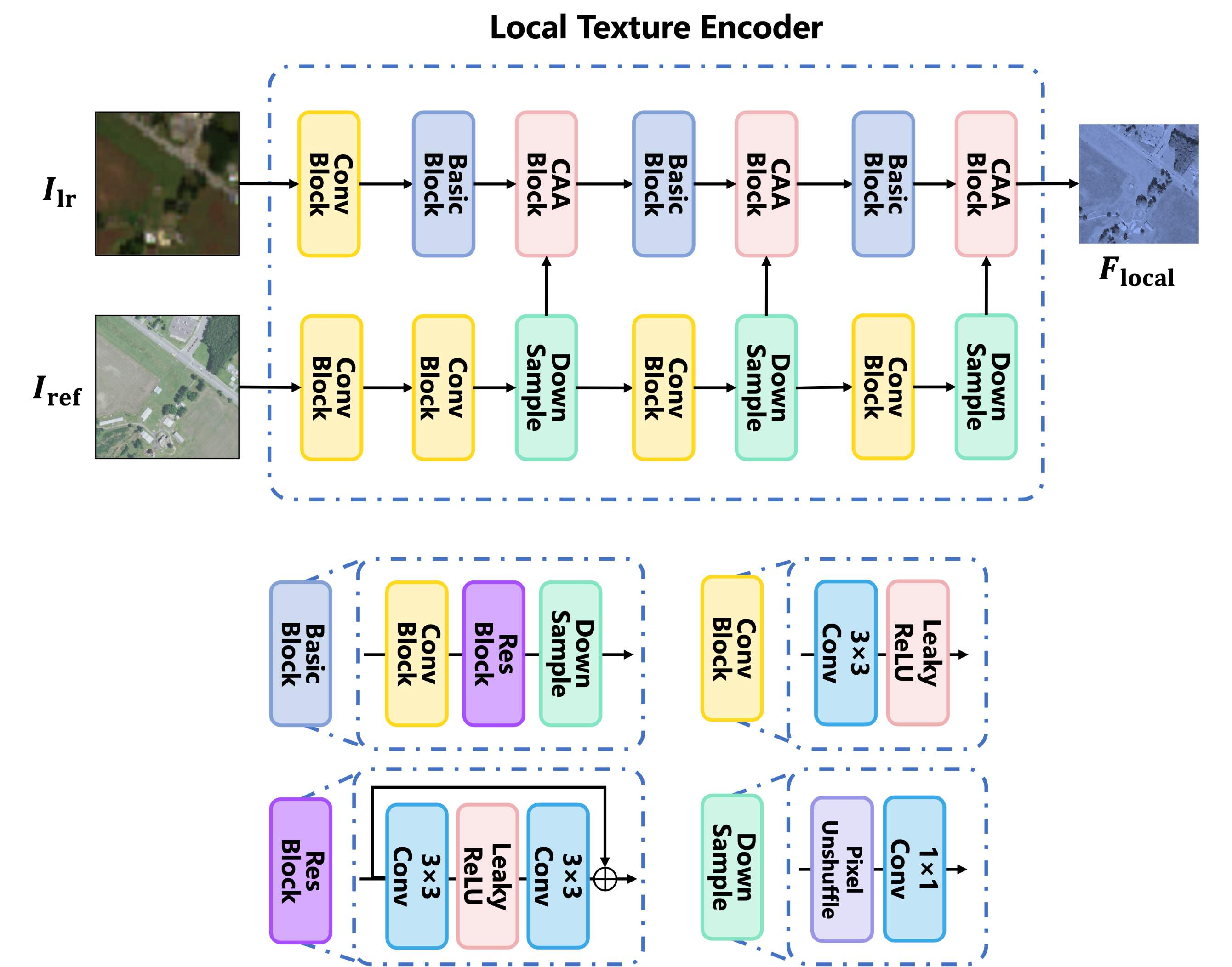}
    \caption{Structure of the Local Texture Encoder. The local texture encoder extracts fine-grained textures from the reference image to enhance the feature representation of the LR image. The Change Aware Attention (CAA) Block is employed to mitigate potential land cover changes while injecting reference information into the LR image features.}
    \label{fig:lte}
\end{figure}
\subsubsection{Local Texture Encoder}
The overall architecture of the local texture encoder is illustrated in Fig. \ref{fig:lte}. It adopts a multi-scale dual-branch design. In the branch of $I_{\mathrm{ref}}$, multiple convolutional and downsampling modules are used to extract multi-scale reference features. In the branch of $I_{\mathrm{lr}}$, several basic modules are employed to extract multi-scale low-resolution features. At each corresponding scale, the features from the two branches are fused through a Change-Aware Attention (CAA) Block, which facilitates the integration of reference information while accounting for potential content changes.
\begin{figure}[htb]
    \centering
    \includegraphics[width=\linewidth]{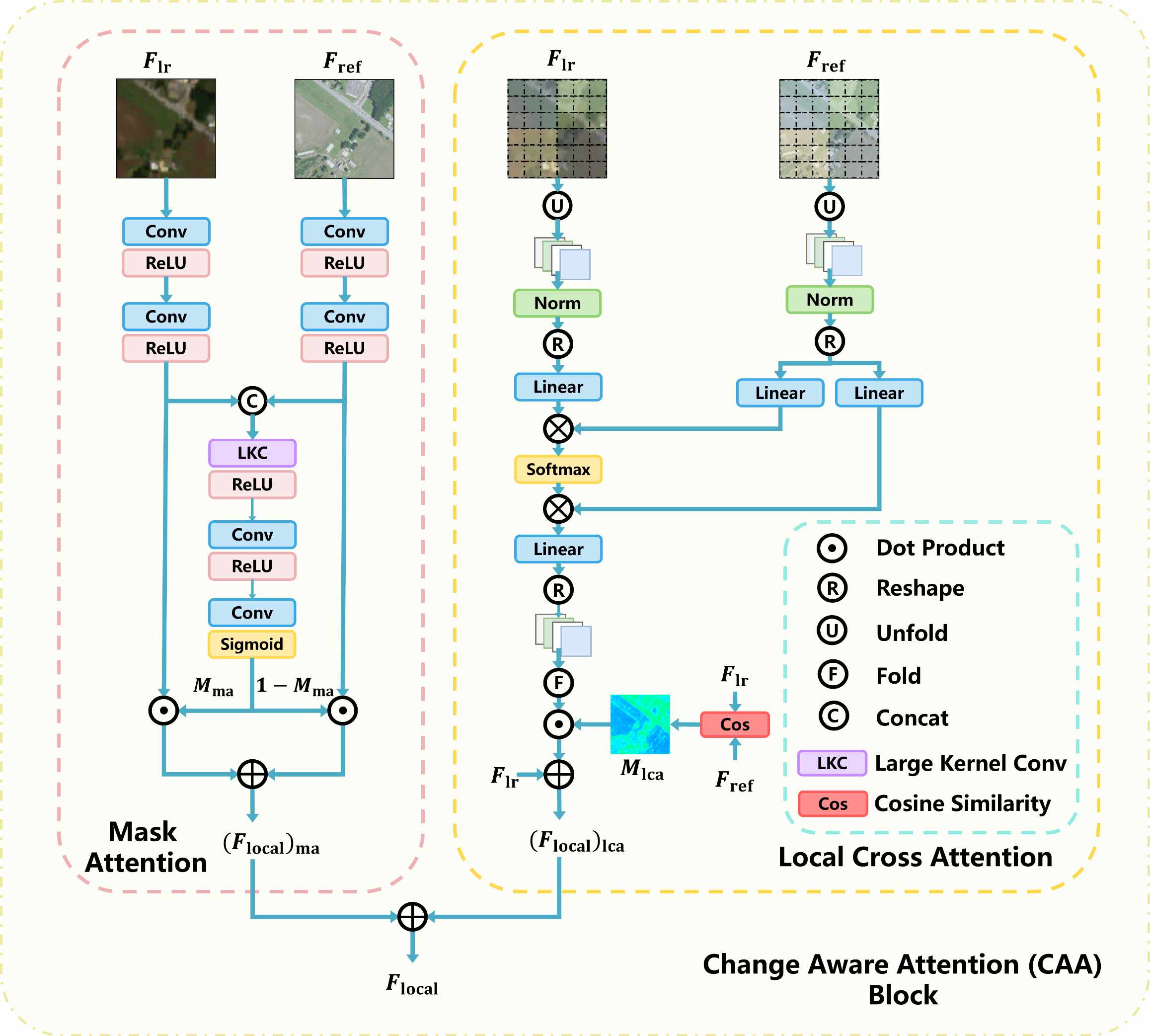}
    \caption{The structure of the Change-Aware Attention Block consists of two sub-modules: a mask attention module and a local cross-attention module, which collaboratively mitigate potential land cover changes and fully exploit the local information from the reference image.}
    \label{fig:lca}
\end{figure}
The structure of the CAA Block is illustrated in Fig. \ref{fig:lca}. It is designed to mitigate potential land cover changes while effectively incorporating locally relevant information from the reference image. The CAA block consists of two sub-modules: a mask attention module and a local cross attention module. The mask attention module learns a spatial attention mask, which is then used to perform a weighted aggregation of the LR and reference features:
\begin{align}
    \hat{F}_\mathrm{lr} &= \mathrm{Conv}\left( F_\mathrm{lr} \right) \notag \\
    \hat{F}_\mathrm{ref} &= \mathrm{Conv}\left( F_\mathrm{ref} \right) \notag \\
    M_{\mathrm{ma}} &= \mathrm{Sigmoid}\left( \mathrm{Conv}\left( \left[ \hat{F}_\mathrm{lr}, \hat{F}_\mathrm{ref} \right] \right) \right) \notag \\
    \left( \mathrm{F}_\mathrm{local} \right)_{\mathrm{ma}} &= M_{\mathrm{ma}} \odot \hat{F}_\mathrm{ref} + \left( 1 - M_{\mathrm{ma}} \right) \odot \hat{F}_\mathrm{lr}
    \label{equ:m_ma}
\end{align}

The local cross-attention module performs fine-grained point-to-point matching between LR and reference features within a restricted attention window. Then it computes an attention map via dot-product similarity and applies it to selectively enhance the output features. First, the input features $F\in \mathbb{R}^{B \times C \times H \times W}$ are passed through an unfold operation, followed by normalization, and then reshaped to the target dimensions $F\in \mathbb{R}^{(B\times \frac{H}{h}\times \frac{W}{w}) \times C \times h \times w}$:
\begin{align}
    \hat{F}_\mathrm{lr} &= \mathrm{Reshape} \left( \mathrm{Norm} \left( \mathrm{Unfold} \left( F_\mathrm{lr} \right) \right) \right) \notag \\
    \hat{F}_\mathrm{ref} &= \mathrm{Reshape} \left( \mathrm{Norm} \left( \mathrm{Unfold} \left( F_\mathrm{ref} \right) \right) \right)
    \label{equ:lca_pre}
\end{align}
Here, $(h,w)$ indicates the spatial dimensions of the local window used for attention computation. This ensures that the subsequent attention computation is restricted to local windows, thereby effectively leveraging the local similarity between the reference and LR images. Subsequently, cross-attention is employed to extract beneficial local information from the reference features to enhance the LR features:
\begin{align}
    Q &= \hat{F}_\mathrm{lr} W_{\mathrm{q}} \notag \\
    K, V &= \hat{F}_\mathrm{ref} W_{\mathrm{k}}, \hat{F}_\mathrm{ref} W_{\mathrm{v}} \notag \\
    F_{\mathrm{ca}} &= \mathrm{Softmax} \left( \frac{Q K^{\mathrm{T}}}{\sqrt{d_k}} \right) V \notag \\
    F_{\mathrm{ca}} &= \mathrm{Fold} \left( \mathrm{Reshape} \left( F_{\mathrm{ca}} \right) \right)
    \label{equ:lca}
\end{align}
Although local cross-attention can adaptively extract useful information within each window, it treats all windows equally and thus fails to handle large-scale land cover changes---especially those beyond the window size. To address this, we first compute the cosine similarity between the features of the LR image and the reference image, and then use it to modulate the residual connections in the cross-attention module:
\begin{align}
    \left( M_{\mathrm{lca}} \right)_{b,h,w} &= 
    \frac{
        \left( F_\mathrm{lr} \right)_{b,:,h,w} \cdot \left( F_\mathrm{ref} \right)_{b,:,h,w}
    }{
        \left\| \left( F_\mathrm{lr} \right)_{b,:,h,w} \right\|_2 \, 
        \left\| \left( F_\mathrm{ref} \right)_{b,:,h,w} \right\|_2
    } 
    \label{equ:m_lca} \\
    \left( \mathrm{F}_\mathrm{local} \right)_{\mathrm{lca}} &= 
    M_{\mathrm{lca}} \odot F_\mathrm{ca} + 
    \left( 1 - M_{\mathrm{lca}} \right) \odot F_\mathrm{lr}
    \label{equ:lca_final}
\end{align}

Finally, the outputs of the two sub-modules are combined via element-wise addition to produce the final output of the CAA module:
\begin{align}
    F_{\mathrm{local}} &= \left( \mathrm{F}_\mathrm{local} \right)_{\mathrm{ma}} + \left( \mathrm{F}_\mathrm{local} \right)_{\mathrm{lca}}
    \label{equ:caa}
\end{align}

\subsubsection{Adapter Design}
After processing by the local texture encoder, a conditional tensor $F_{\mathrm{local}}$ enriched with high-frequency details is obtained. Motivated by findings from recent studies in AIGC \citep{zhang2023adding, mou2024t2i}, we adopt the adapter architecture to guide the generation of the SD model. The adapter is a multi-scale module, where each scale consists of several residual blocks followed by a downsampling module. It takes the conditional tensor $F_{\mathrm{local}}$ as input and extracts multi-scale features in a coarse-to-fine manner, producing features at multiple spatial resolutions:
\begin{gather}
    \{F_{c}^{1}, F_{c}^{2}, ... ,F_{c}^{K}\} = \mathrm{Adapter}\left(F_{\mathrm{local}}\right)
    \label{equ:adapter}
\end{gather}
Subsequently, these features are element-wisely added to different layers of the denoising UNet. Considering that image SR task places a higher emphasis on fidelity, we employ conditional control over both the encoding and decoding layers of the denoising U-Net \citep{mou2024t2i}:
\begin{align}
    F_{\mathrm{enc}}^{i} &= F_{\mathrm{enc}}^{i} + F_{c}^{i}, \ \ i \in \{1, 2, ..., K\} \notag \\
    F_{\mathrm{dec}}^{i} &= F_{\mathrm{dec}}^{i} + F_{c}^{i}, \ \ i \in \{1, 2, ..., K\}
    \label{equ:guide_decoder}
\end{align}
Here, $F_{\mathrm{enc}}^{i}$ and $F_{\mathrm{dec}}^{i}$ represent the features of the denoising UNet's encoding and decoding layers, respectively. In this way, the information from the LR and reference images can be effectively aligned with the inherent priors of the pre-trained SD model, thereby guiding its denoising process. 

\subsection{Global Fusion Branch}
\begin{figure}[htb]
    \centering
    \includegraphics[width=\linewidth]{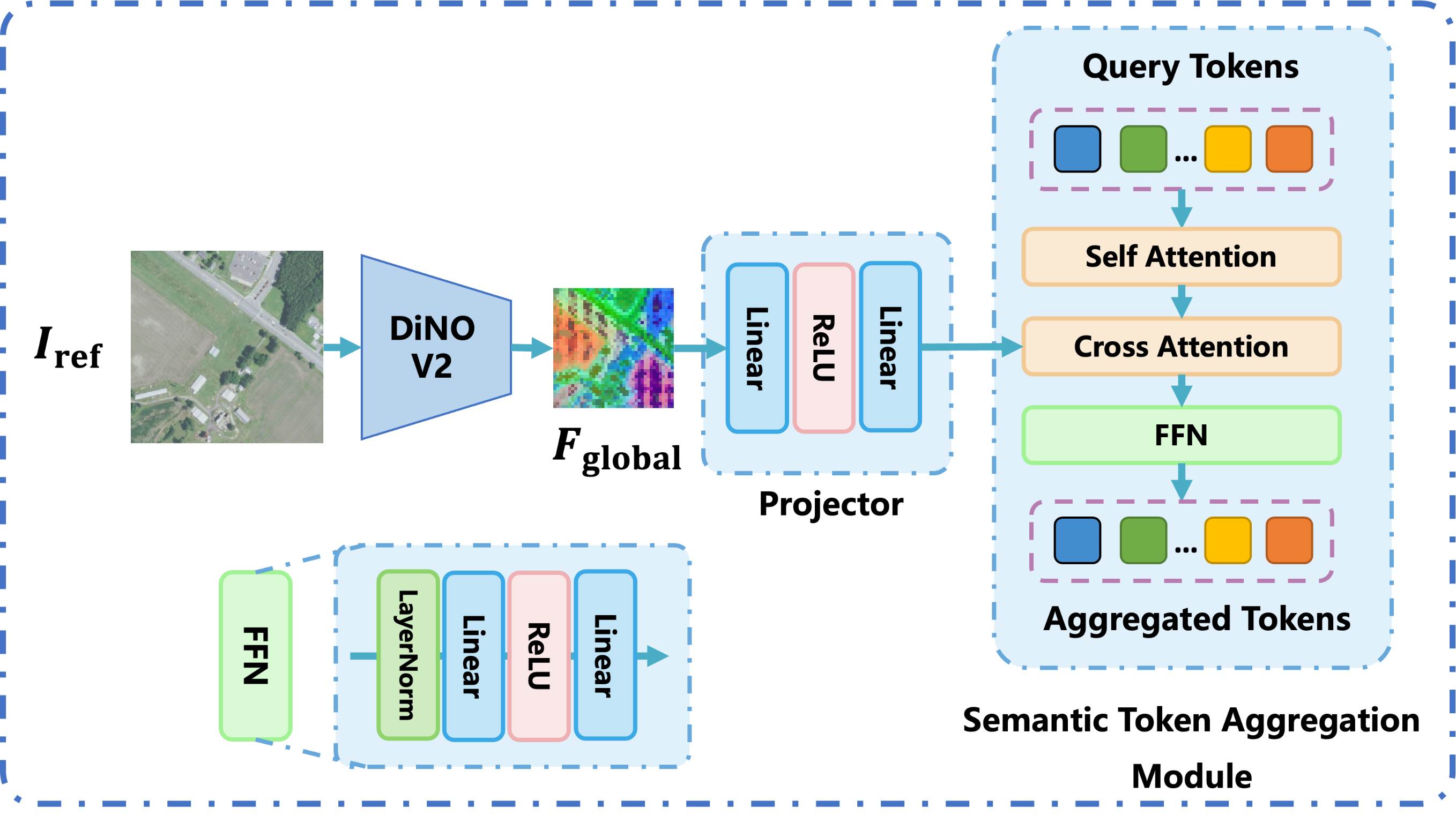}
    \caption{Structure of the global fusion branch. We adopt DiNOv2 in place of CLIP to extract foundation features from the reference image, which are subsequently refined through a semantic token aggregation module.}
    \label{fig:sta}
\end{figure}

The local fusion branch is limited to leveraging only local information from the reference image and fails to effectively capture its global contextual cues. To address this limitation, we introduce a global fusion branch, whose architecture is illustrated in Fig. \ref{fig:sta}. Recent studies \citep{jiang2023clip} have shown that DiNOv2 \citep{oquab2023dinov2}, a self-supervised vision encoder, is capable of capturing finer-grained visual information compared to CLIP. However, the token representations produced by DiNOv2 are numerous and exhibit significant spatial redundancy. To address this, we propose a semantic token aggregation module, inspired by BLIP-2 \citep{li2023blip}, to distill the original features into a compact set of semantic tokens. The aggregated semantic tokens are then used to replace the text embeddings in Stable Diffusion and are injected into the UNet via cross-attention.

Specifically, we first extract global features from the reference image using a pretrained DiNOv2 model. These features are then reshaped and projected through two linear layers:
\begin{align}
    F_{\mathrm{global}} &= f_\mathrm{dinov2}\left(F_{\mathrm{ref}}\right) \notag \\
    F_{\mathrm{proj}} &= f_{\mathrm{proj}}\left(F_{\mathrm{global}}\right)
    \label{equ:sta_pre}
\end{align}
Here, $f_\mathrm{dinov2}$ and $f_{\mathrm{proj}}$ denote the DiNOv2 encoder and the projection layers, respectively. Then a set of learnable queries $\mathcal{Q}$ is initialized. These queries are first enhanced through a self-attention mechanism to strengthen their mutual interactions. Subsequently, they are used to perform cross-attention with the projected features $F_{\mathrm{proj}}$, allowing each query to retrieve the most relevant visual information from the reference features. Finally, these features are passed through a feed-forward network to produce the aggregated semantic tokens $F_{\mathrm{aggr}}$. This process can be formulated as follows:
\begin{align}
    F_{\mathrm{aggr}} &= f_\mathrm{ffn}\left(f_\mathrm{cross\text{-}attn}
    \left(f_\mathrm{self\text{-}attn}\left(\mathcal{Q}\right), F_{\mathrm{proj}}\right)\right)
    \label{equ:sta}
\end{align}
Here, $f_\mathrm{self\text{-}attn}$, $f_\mathrm{cross\text{-}attn}$, and $f_\mathrm{ffn}$ denote the self-attention, cross-attention, and feed-forward network, respectively. These operations aggregate the original global features $F_{\mathrm{global}} \in \mathbb{R}^{M \times C}$ into a set of semantic tokens $F_{\mathrm{aggr}} \in \mathbb{R}^{N \times C}$ (with $M \gg N$) , significantly reducing the token count and thereby lowering subsequent computational costs. The aggregated tokens $F_{\mathrm{aggr}}$ are then injected into the UNet via cross-attention to provide semantic guidance during the denoising process.

\subsection{Augmentation Strategy}
\begin{figure}[htb]
    \centering
    \includegraphics[width=\linewidth]{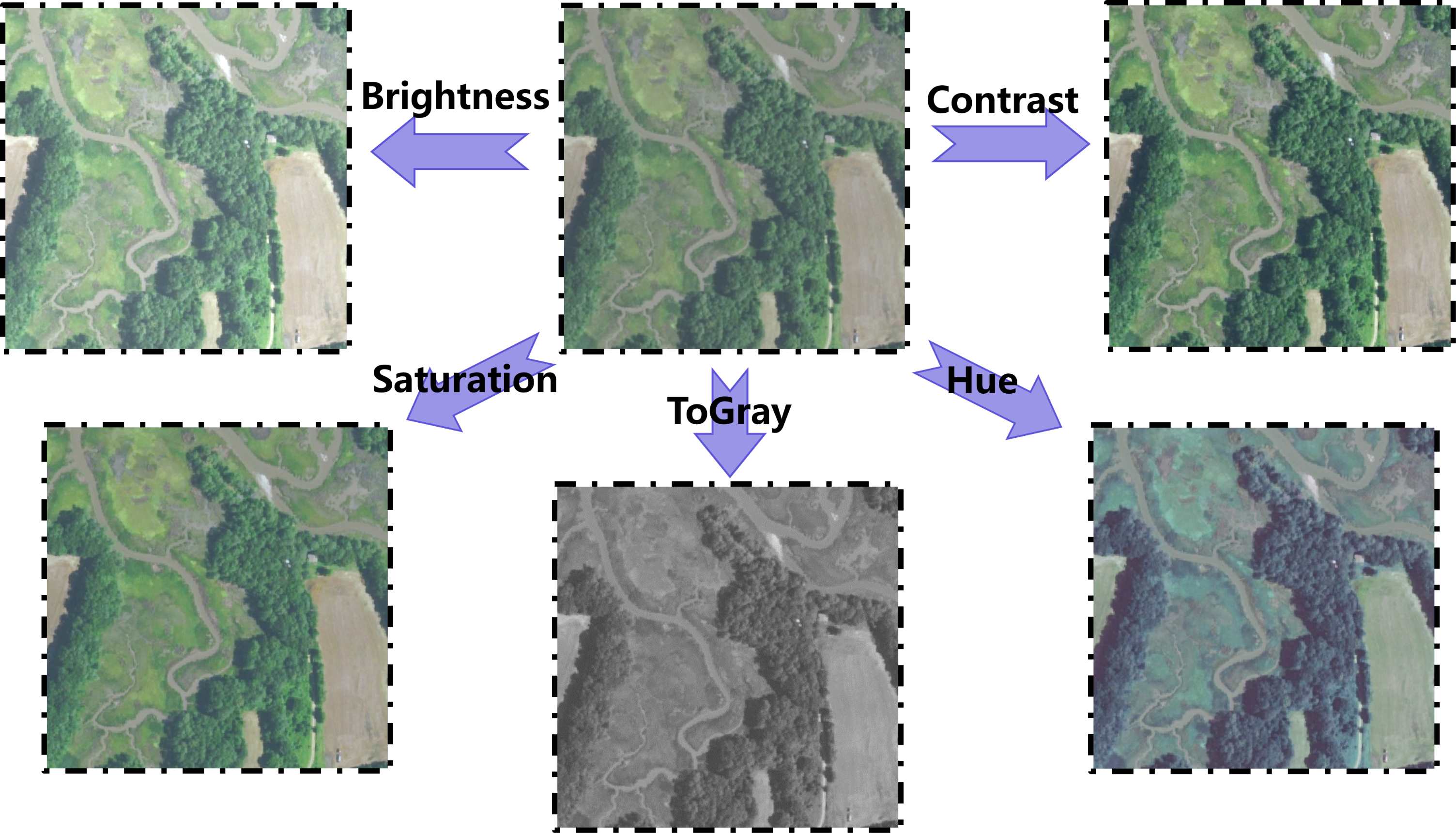}
    \caption{Illustration of the reference image augmentation strategy. During training, we randomly perturb the style attributes of the reference image.}
    \label{fig:aug_ref}
\end{figure}

Due to differences in imaging time, the reference and HR images often share similar content but differ in style. To encourage the model to focus on content rather than unreliable style cues, we apply random style perturbations to the reference image during training by modifying brightness, contrast, saturation, hue, and occasionally converting it to grayscale (as shown in Fig. \ref{fig:aug_ref}). These stochastic augmentations simulate diverse visual conditions, guiding the model to attend to invariant structural and semantic features. This serves as an implicit regularization strategy, improving robustness to style variations and enhancing generalization. It also aligns with the contrastive learning principle \citep{chen2020simple} that stronger augmentations promote more discriminative and transferable representations.

\subsection{Optimization and Inference}
During training, we optimize the model using an L2 loss in the latent space \citep{ho2020denoising, rombach2022high}:
\begin{equation}
\mathcal{L}_{\text{diff}} = \mathbb{E}_{I_{\mathrm{hr}}, I_{\mathrm{lr}}, I_{\mathrm{ref}}, t, \epsilon}  \| \epsilon - \epsilon_\theta(z_t, t, I_{\mathrm{lr}}, I_{\mathrm{ref}}) \|^2.
\end{equation}
As illustrated in Fig. \ref{fig:arch}, the modules involved in training include the local texture encoder, the projector, the STA module, and all attention layers within the denoising UNet.

Considering that the vanilla diffusion model requires an iterative sampling process during inference, which leads to high computational costs, we propose a Better Start strategy to reduce the number of inference steps. Inspired by \citep{wu2024seesr}, the LR image can serve as a starting point for the reverse denoising process, thereby shortening the denoising trajectory and lowering the required number of steps: 
\begin{align}
    I_{\mathrm{start}} &= I_{\mathrm{lr}}, z_{0}^{\mathrm{start}} = \mathcal{E}(I_{\mathrm{start}}) \\
    z_{t'}^{\mathrm{start}} &= \sqrt{\bar{\alpha}_t}z_{0}^{\mathrm{start}}+\sqrt{1 - \bar{\alpha}_t}\epsilon
    \label{equ:better_start}
\end{align}
This enables us to replace $z_{T}$ with $z_{t'}^{\mathrm{start}}$ as the initial state of the reverse denoising process, thereby reducing the number of required denoising steps from $T$ to $t'$. However, for real-world satellite imagery, the LR image often differs significantly from the GT image in both spatial resolution and visual style. Directly using the LR image as the starting point significantly degrades the inference quality. Therefore, we train a lightweight SR model and perform an initial fusion of the LR and reference images. The resulting preliminary fusion is then used as the starting point for denoising, which effectively reduces the number of inference steps while maintaining satisfactory performance.

Additionally, to fundamentally mitigate the model's over-reliance on the reference image, we propose a reference strength control strategy. It allows users to provide a scalar factor $s$ or a spatial mask $m$ during inference to control the amount of reference information injected into the model. For the local branch, this is achieved by modifying the two spatial attention maps ($M_{\mathrm{ma}}$ in Eq. (\ref{equ:m_ma}) and $M_{\mathrm{lca}}$ in Eq. (\ref{equ:m_lca})) within the change aware attention block. The user-provided factor $s$ is used to scale the attention maps and the mask $m$ is used to filter them:
\begin{align}
    M_{\mathrm{ma}} \leftarrow s \cdot m \odot M_{\mathrm{ma}} \notag \\
    M_{\mathrm{lca}} \leftarrow s \cdot m \odot M_{\mathrm{lca}}
    \label{equ:streng_control}
\end{align}
For the global branch, this can be achieved by modifying the cross-attention mechanism. Specifically, we use the factor $s$ to modulate the residual connection of the cross-attention $f_\mathrm{cross\text{-}attn}$, and apply the mask $m$ as the attention mask to control the influence of the reference information:
\begin{align}
    F_{\mathrm{cross-attn}} = Q + s \cdot \mathrm{Attn}\left(Q, K, V, m\right)
    \label{equ:streng_control_global}
\end{align}
Here, $\mathrm{Attn}$ denotes the masked attention mechanism, which can be formulated as follows:
\begin{align}
\text{Attn}(Q, K, V, m) = \mathrm{Softmax}\left( \frac{QK^\top}{\sqrt{d_k}} + \log\left(m\right) \right) V
\end{align}
In this way, setting certain regions in $m$ to zero forces their corresponding cross-attention scores to negative infinity, thereby excluding those regions from the attention output. During training, we randomly set $s$ to zero with a certain probability $p$, forcing the model to rely solely on the LR image for reconstruction. This enables the model to modulate the amount of reference information injected during inference via $s$ and $m$. 

\begin{figure*}[htb]
    \centering
    \includegraphics[width=\linewidth]{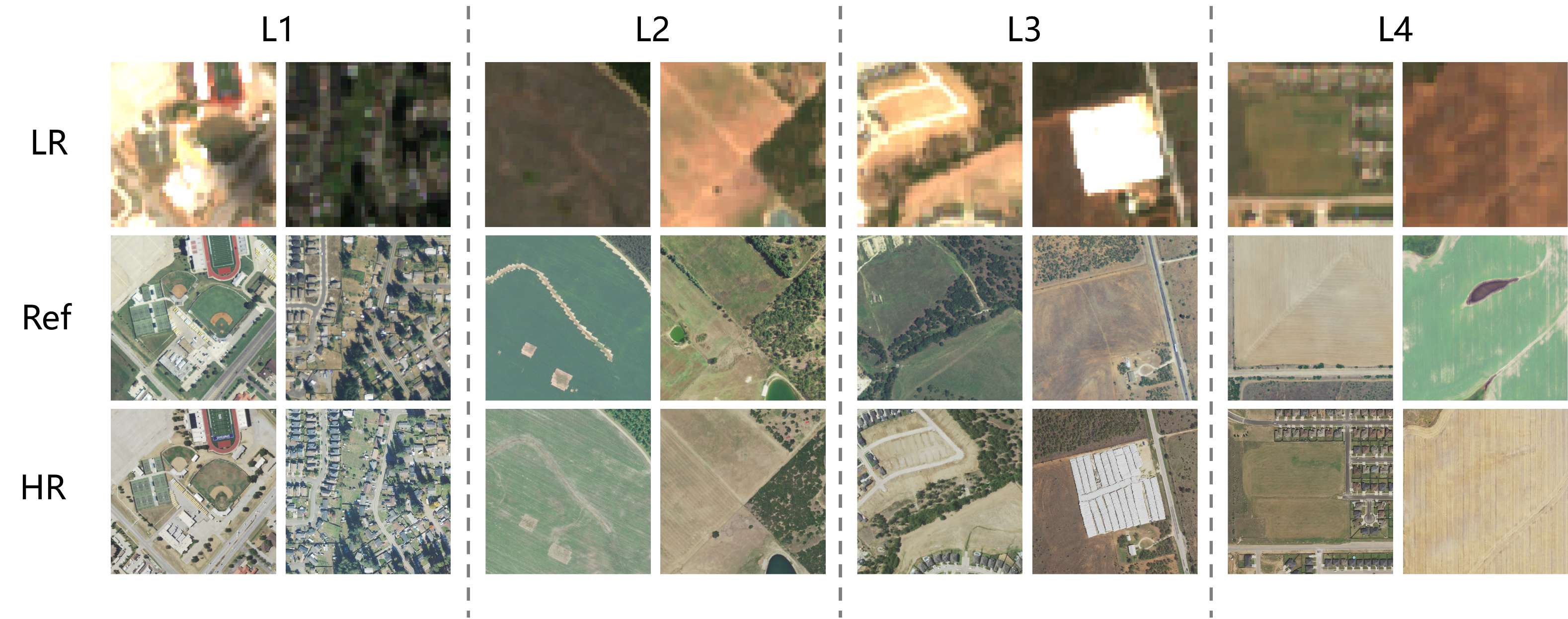}
    \caption{An example of a Ref-LR-HR data triplet from Real-RefRSSRD. The dataset encompasses a diverse range of land cover types and various degrees of land cover changes.}
    \label{fig:real_refsr_dataset}
\end{figure*}
\section{Experiments and discussion}
\subsection{Dataset}
As far as we know, there is currently a lack of real-world reference-based super-resolution datasets for remote sensing imagery. To facilitate further research on this task, we construct a real-world remote sensing RefSR dataset, named Real-RefRSSRD. This dataset is divided into training, validation, and test subsets, and covers a diverse range of scenes, including residential areas, industrial zones, farmland, forests, water bodies, playgrounds, and roads (as shown in Fig. \ref{fig:real_refsr_dataset}).

\begin{figure}[htb]
    \centering
    \includegraphics[width=\linewidth]{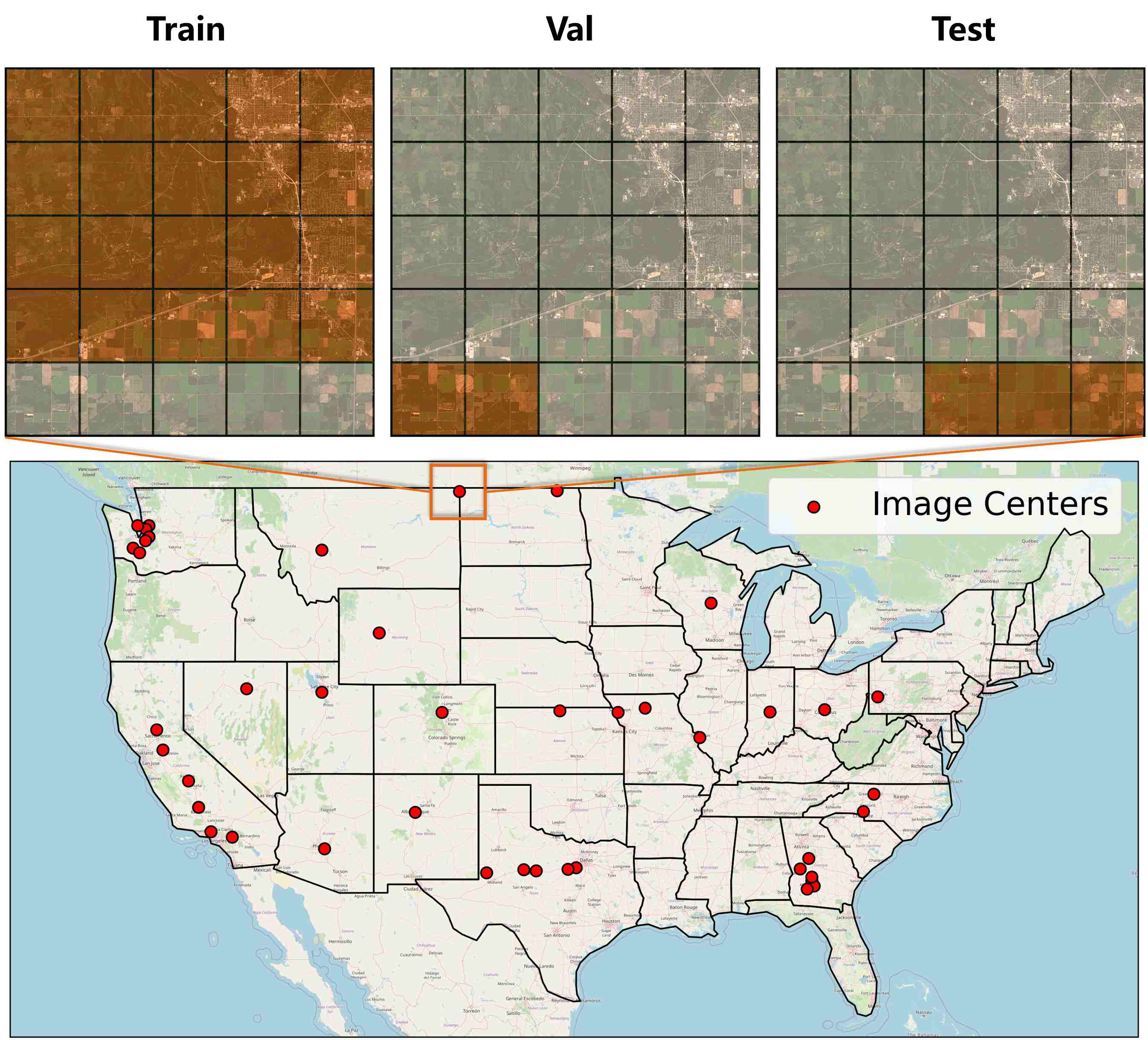}
    \caption{The spatial distribution of the sampled locations and the strategy for splitting the dataset into training, validation, and test sets.}
    \label{fig:sample_points}
\end{figure}

\begin{figure}[htb]
    \centering
    \includegraphics[width=\linewidth]{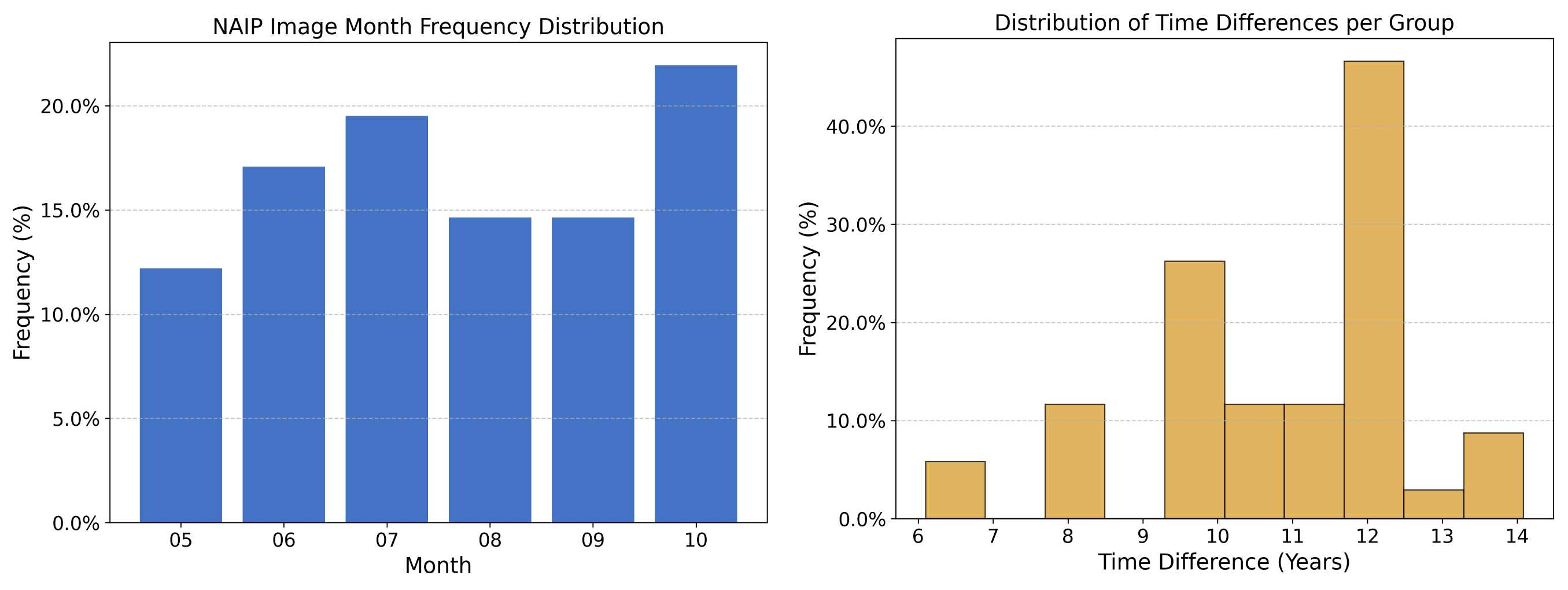}
    \caption{Distribution over months the images were captured and time gaps between reference and HR images.}
    \label{fig:time_dist}
\end{figure}

\begin{table}[htb]
\centering
\caption{Detailed information about Real-RefRSSRD}
\begin{tabular}{ll}
\toprule
\textbf{Attribute} & \textbf{Value} \\
\midrule
Train/Val/Test Split & 74093/172/1456 \\
Image Size (HR/Ref/LR) & 480 px/480 px/48 px   \\
GSD (HR/Ref/LR) & 1 m/1 m/10 m   \\
HR Image Source & NAIP (2020--2023) \\
Ref Image Source & NAIP (2009--2015) \\
LR Image Source & Sentinel-2 (2020--2023) \\
\bottomrule
\end{tabular}
\label{tab:dataset}
\end{table}
Detailed information on Real-RefRSSRD is presented in Table \ref{tab:dataset}. We first collect recent observations from the NAIP satellite as HR images. Each HR image is then paired with a temporally and spatially aligned Sentinel-2 observation (with a time gap of less than 7 days and with cloud coverage below 10\%), which serves as the LR image. Finally, we use historical NAIP observations, which are spatially aligned with the HR image but exhibit a significant temporal gap (ranging from 6 to 14 years), as the reference images. All NAIP images have a ground sampling distance (GSD) of 1 m, while the Sentinel-2 images have a GSD of 10 m. In total, we collect 43 scenes, which are distributed across various regions of the United States (as shown in Fig. \ref{fig:sample_points}). Considering that urban areas contain more complex and challenging land cover patterns, we focus our sampling more heavily on metropolitan regions while reducing the sampling frequency in mountainous areas of the central United States. Furthermore, we analyze the temporal distribution of the dataset, as illustrated in Fig. \ref{fig:time_dist}. The NAIP images are acquired primarily between May and October, and our dataset provides relatively uniform coverage across these months. To increase the difficulty and realism of the dataset, we also ensure that the reference and HR images exhibit substantial temporal gaps, mostly exceeding 10 years.

Finally, all NAIP/Sentinel-2 images are cropped into 480 px $\times$ 480 px/48 px $\times$ 48 px patches and divided into training, validation, and test sets. A reasonable partition of training, validation, and test sets is essential to ensure the generalizability of machine learning models. In the context of remote sensing imagery, spatial correlations may lead to leakage between train and test stages. To mitigate this issue, we adopt a spatially independent partitioning strategy within each scene, following a similar approach to \citep{razzak2023multi}, as illustrated in Fig. \ref{fig:sample_points}. Specifically, the top portion of each scene is designated for training, while the lower-left and lower-right regions are used for validation and testing, respectively. This design effectively eliminates the risk of data leakage.

In real-world scenarios, the provided reference images may exhibit varying degrees of similarity to the HR images. To evaluate the generalization capability of our algorithm, we use the LPIPS metric between the reference and HR images as a measure of similarity, and divide the test set into four subsets---L1, L2, L3, and L4---with progressively decreasing levels of similarity (as shown in Fig. \ref{fig:real_refsr_dataset}). Each subset contains 200 image pairs. In the subsequent experiments, unless otherwise specified, all reported results are evaluated on the entire test set.

\subsection{Implementation Details}
Our proposed CRefDiff is built upon Stable Diffusion 2.1-base. All experiments are conducted using the PyTorch framework on a single NVIDIA GeForce RTX 3090 GPU. During training, only the local texture encoder, adapter, projector, semantic token aggregation module, and all attention layers in the UNet are optimized, resulting in a total of 170M trainable parameters. The model is trained for 400K steps with a batch size of 8, using the AdamW optimizer with a learning rate of $5\times10^{-5}$. During training, we randomly perturb the style attributes of the reference image, including adjustments to contrast, saturation, hue, brightness, and even conversion to grayscale. Additionally, we occasionally drop the reference image entirely or use the HR image as the reference. These strategies are designed to improve the model’s ability to flexibly leverage reference information under varying degrees of similarity.

For comparison, we select four SISR methods and four RefSR methods. The SISR baselines include TTST \citep{xiao2024ttst}, ESRGAN \citep{wang2018esrgan}, EDiffSR \citep{xiao2023ediffsr}, and DiffBIR \citep{lin2023diffbir}, while the RefSR baselines include TTSR \citep{zhang2020texture}, DATSR \citep{cao2022reference}, RRSGAN \citep{dong2021rrsgan}, and RGTGAN \citep{tu2024rgtgan}. All methods are trained on the training set of Real-RefRSSRD using the official code and configurations provided in their papers.

\subsection{Metrics}
To comprehensively evaluate the various SR methods, we employ both full-reference and no-reference image quality metrics. To assess pixel-level fidelity, we use the peak signal-to-noise ratio (PSNR) and the structural similarity index (SSIM) \citep{wang2004image}, both calculated on the Y channel of the YCbCr color space. To evaluate perceptual fidelity, we adopt learned perceptual image patch similarity (LPIPS) \citep{zhang2018unreasonable} and deep image structure and texture similarity (DISTS) \citep{ding2020image}. Furthermore, Fréchet Inception Distance (FID) \citep{heusel2017gans} and the patch Fréchet Inception Distance (pFID) \citep{chai2022any} are used to gauge the distributional disparity between the ground truth images and the SR images. The pFID is a modified version of FID that computes the score after applying random resizing and cropping to the images, making it more sensitive to blurring or artifacts at high-resolution levels. In addition, we incorporate two no-reference metrics MUSIQ \citep{ke2021musiq} and CLIPIQA \citep{wang2023exploring} to further assess the overall effectiveness of each method. More importantly, SR for RS not merely aims to improve visual quality the the reconstructed SR images, the results also should be applicable to the downstream remote sensing image interpretation tasks. Thus, we also compare the SR images generated by the proposed CRefDiff and that of the comparison methods on the scene recognition and semantic segmentation tasks using Accuracy and mIoU, respectively.

\subsection{Comparison With State-of-the-Arts}
\begin{table*}[htp]
\centering
\caption{Results of various methods on the Real-RefRSSRD test set. For each metric, $\uparrow$ and $\downarrow$ indicate that higher or lower values correspond to better performance, respectively. The bold and underlined numbers denote the best and second-best results, respectively.}
\label{tab:comparison}

\resizebox{\textwidth}{!}{
\begin{tabularx}{\textwidth}{@{}c c >{\centering\arraybackslash}X >{\centering\arraybackslash}X >{\centering\arraybackslash}X >{\centering\arraybackslash}X >{\centering\arraybackslash}X >{\centering\arraybackslash}X >{\centering\arraybackslash}X >{\centering\arraybackslash}X@{}}
\toprule
 & Method/Metric & PSNR$\ \uparrow$ & SSIM$\ \uparrow$ & LPIPS$\ \downarrow$ & DISTS$\ \downarrow$ & FID$\ \downarrow$ & pFID$\ \downarrow$  & MUSIQ$\ \uparrow$ & CLIPIQA$\ \uparrow$ \\
\midrule
\multirow{4}{*}{SISR} & TTST & 20.55 & 0.4585 & 0.8624 & 0.5049 & 200.7 & 248.6 & 17.45 & 0.2371 \\
 & ESRGAN & 18.46 & 0.3603 & 0.4509 & 0.2670 & 70.41 & 48.23 & 52.21 & 0.5422 \\
 & EDiffSR & 18.10 & 0.3492 & 0.5181 & 0.3251 & 168.5 & 128.0 & 47.82 & \underline{0.5626} \\
 & DiffBIR & 19.08 & 0.4084 & 0.4741 & 0.2642 & \underline{54.54} & \underline{36.88} & \underline{52.95} & 0.5489 \\
\midrule
\multirow{5}{*}{RefSR} & TTSR & 20.75 & 0.4259 & \underline{0.4366} & \underline{0.2631} & 79.48 & 55.18 & 48.87 & 0.4441 \\
 & DATSR & \textbf{21.53} & \textbf{0.4901} & 0.4706 & 0.2861 & 85.08 & 71.09 & 32.80 & 0.3611 \\
 & RRSGAN & 20.71 & 0.4422 & 0.4659 & 0.2998 & 100.9 & 79.64 & 43.05 & 0.4752 \\
 & RGTGAN & 20.67 & 0.4378 & 0.4802 & 0.2858 & 56.09 & 39.88 & 48.87 & 0.4459 \\
 & Ours & \underline{21.33} & \underline{0.4691} & \textbf{0.3852} & \textbf{0.2304} & \textbf{46.59} & \textbf{30.54} & \textbf{54.44} & \textbf{0.6014} \\
\bottomrule
\end{tabularx}
}
\end{table*}

\subsubsection{Quantitative Comparison}
Table \ref{tab:comparison} provides a comprehensive comparison of all methods across the entire test set. It can be observed that our method achieves the best performance across all perceptual metrics, while also attaining second-best results in pixel-level fidelity metrics such as PSNR and SSIM. Among SISR methods, TTST, as a regression-based approach, achieves relatively high SSIM. However, due to the tendency of regression-based methods to produce overly smooth results, its perceptual quality lags significantly behind other methods. In contrast, ESRGAN, EDiffSR, and DiffBIR achieve superior performance in LPIPS, CLIPIQA, and FID, respectively. Owing to the availability of high-frequency details provided by the reference images, RefSR methods significantly outperform SISR methods across most evaluation metrics. DATSR focuses more on pixel-level reconstruction, thus achieving the highest PSNR and SSIM scores. In contrast, TTSR emphasizes the generation of fine-grained textures, resulting in better LPIPS and DISTS performance. Our method effectively leverages both local details and global contextual information from the reference image, striking a balance between fidelity-oriented and perception-oriented metrics. Furthermore, benefiting from the strong generative capacity of the pretrained Stable Diffusion model, our approach achieves a significant advantage in perceptual quality.

\subsubsection{Qualitative Comparison}
\begin{figure*}[t!]
    \centering
    \includegraphics[width=\linewidth]{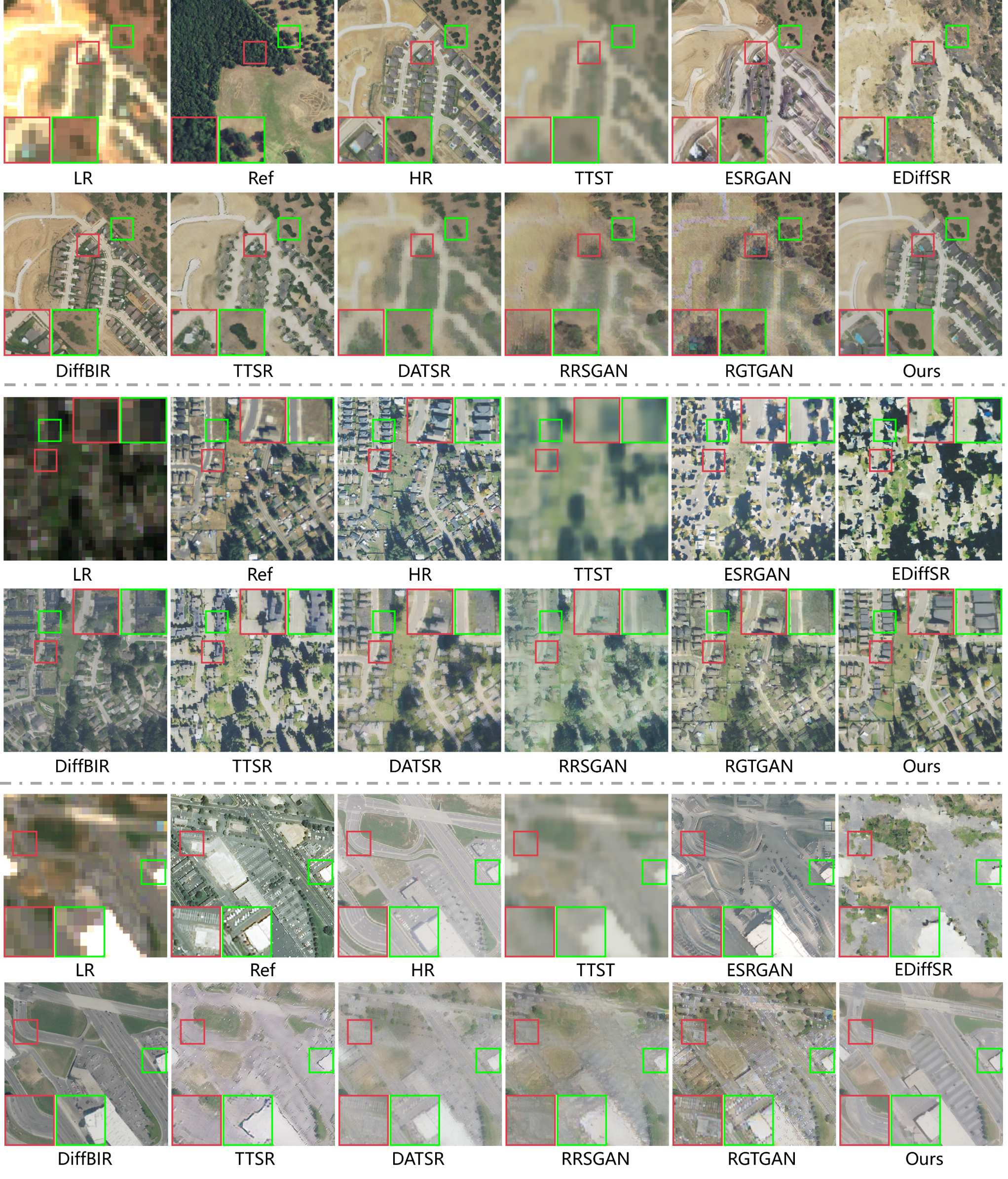}
    \caption{Visual comparison of our method with different SR methods on the test sets.}
    \label{fig:visual_compare}
\end{figure*}

Fig. \ref{fig:visual_compare} presents the qualitative comparison between our method and other SISR and RefSR methods. The first row illustrates a challenging case where substantial land cover changes exist between the reference and HR images. The second row depicts a more common scenario, where only partial changes occur. In both cases, the comparison methods either reconstruct incorrect structures (e.g., EDiffSR, DiffBIR, TTSR) or produce overly blurred results (e.g., TTST, DATSR, RRSGAN). In contrast, our method effectively integrates spatial contextual information from both the reference and LR images and benefits from the strong generative prior of the pretrained Stable Diffusion model. As a result, it successfully reconstructs accurate structures and rich high-frequency details. The final group of examples presents a special case in which certain prominent regions in the reference image have undergone land cover changes (as indicated by the red box), while other prominent regions remain unchanged (as shown in the green box). In this scenario, some RefSR methods suffer from a ``copy-paste'' effect (e.g., DATSR, RGTGAN), directly transferring outdated structures from the reference image. In contrast, our method effectively distinguishes land cover changes by jointly reasoning over the reference and LR images, thereby avoiding such artifacts.
\subsubsection{Downstream Tasks' Performance Comparison}
\begin{figure*}[htb]
    \centering
    \includegraphics[width=\textwidth]{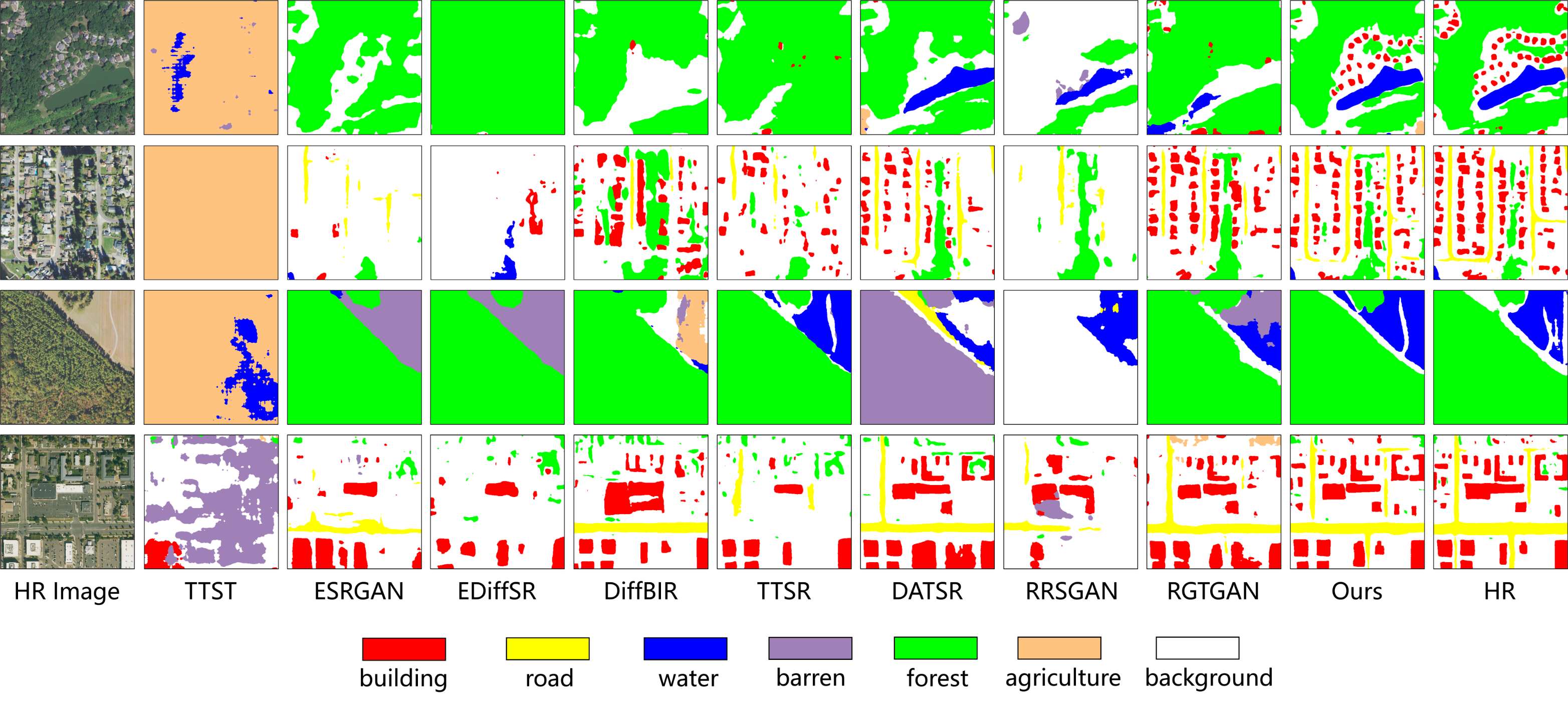}
    \caption{Visual comparison of semantic segmentation results produced by different methods.}
    \label{fig:seg_compare}
\end{figure*}

\begin{figure}[htb]
    \centering
    \includegraphics[width=\linewidth]{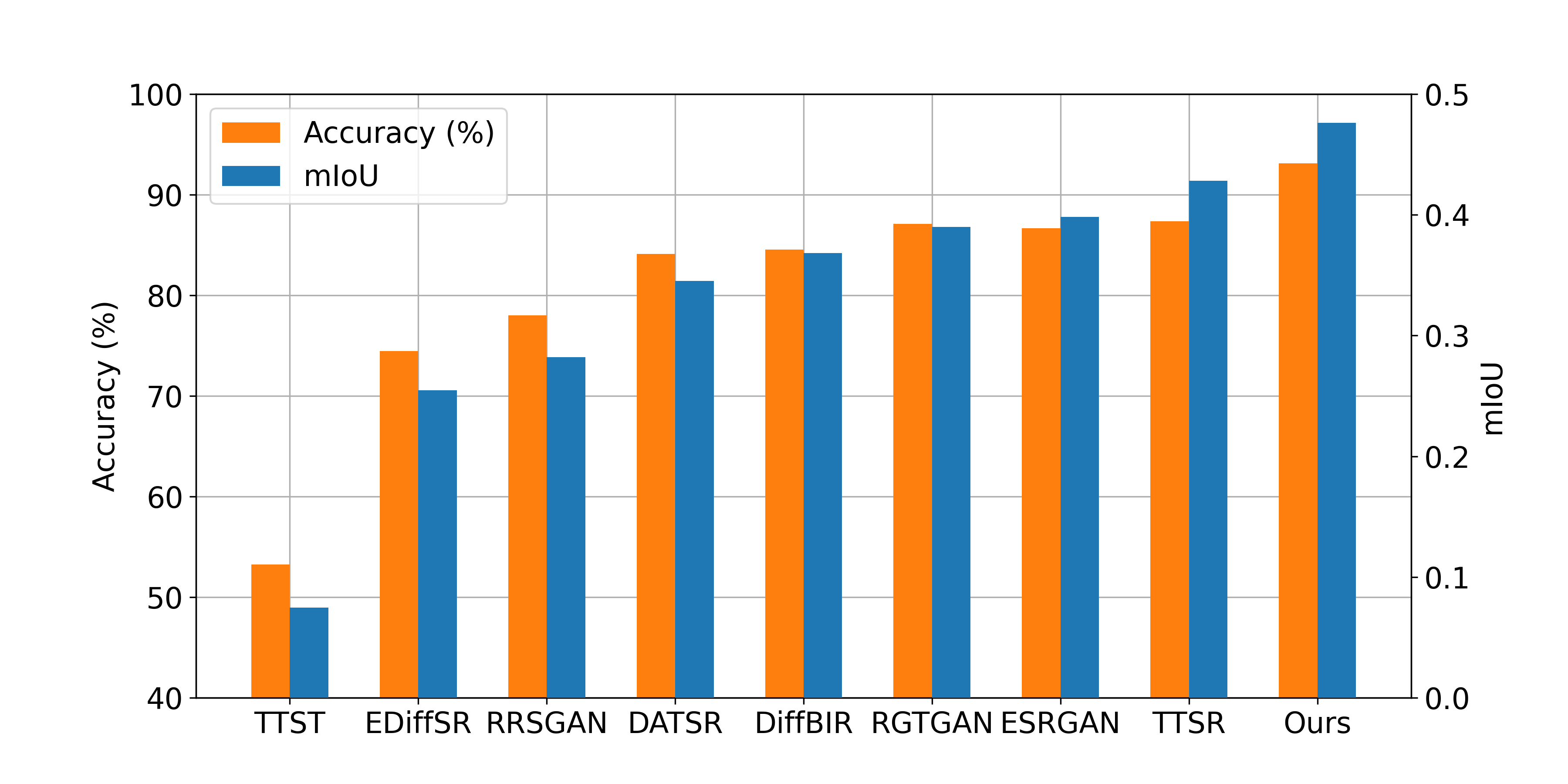}
    \caption{Comparison of scene classification and semantic segmentation accuracy across different SR methods.}
    \label{fig:downstream_task}
\end{figure}

For natural images, visual quality plays a critical role in performance evaluation. However, remote sensing images are not only used for visual interpretation but are also widely applied in downstream machine vision tasks. Therefore, to provide a more comprehensive assessment of SR algorithms, we incorporate two representative tasks, scene classification and semantic segmentation, to evaluate the utility of reconstructed images in machine vision scenarios. Specifically, we employ a pretrained remote sensing foundation model (RVSA \citep{wang2022advancing}) to analyze both the SR results and the original HR images, treating the latter as ground truth labels to compute performance metrics for each SR method. The performance comparison of the two tasks is illustrated in Fig. \ref{fig:downstream_task}. Our method achieves the highest accuracy in the scene classification task, surpassing 90\% in top-5 accuracy. This indicates that the reconstructed images exhibit the strongest global semantic consistency with the original HR images. Similarly, in the semantic segmentation task, our method attains the highest mean Intersection-over-Union (mIoU), demonstrating superior semantic consistency in local details. Among all compared methods, the regression-based method (TTST) shows significantly lower performance than generative approaches, as its overly smooth outputs hinder effective feature extraction and analysis of the pretrained foundation model. In Fig. \ref{fig:seg_compare}, we present the semantic segmentation results of different SR methods. It is evident that our method enables more accurate identification of fine-grained objects such as buildings and roads. In contrast, the comparison methods often suffer from missed detections and misclassifications of land cover objects.

\subsection{Ablation Studies}
\subsubsection{Effectiveness of Global Branch}
The global branch is designed to extract contextual information from the reference image and guide the generation process of the SD model via cross-attention. This mechanism resembles the patch matching strategy in traditional RefSR methods, where relevant regions in the reference image are globally aggregated to support the enhancement of the LR input. Our global branch consists of two components: a DiNO-based feature extractor and a semantic token aggregation module. To validate their effectiveness, we design the following two ablation schemes.

First, we adopt a learnable embedding as the conditioning input for cross-attention and train the model from scratch, a strategy that is commonly used in other SD-based super-resolution methods \citep{wang2023exploiting, lin2023diffbir}. Similar to previous works \citep{sun2024coser, lin2023diffbir, wang2023exploiting}, we select a representative set of evaluation metrics for the ablation study: the fidelity metric PSNR, the perceptual metric LPIPS, and the no-reference metric CLIPIQA. As shown in the first row of Table \ref{tab:ablation_study}, this configuration results in performance degradation across all evaluation metrics. This decline stems from the inability of fixed empty prompts to adapt to varying LR inputs, as well as their limited capacity to convey sufficient contextual information compared to reference image. 
\begin{table*}[htb]
	\caption{Ablation study of our method. Bold represents the best and underline represents the second-best.}
	\label{tab:ablation_study}
	\centering
	\begin{tabularx}{\textwidth}{CCCCCCCCC}
		\hline
         \makecell[c]{Local} &\makecell[c]{Global} &\makecell[c]{SD prior} &PSNR$\ \uparrow$ &SSIM$\ \uparrow$ &LPIPS$\ \downarrow$ &DISTS$\ \downarrow$ &MUSIQ$\ \uparrow$ &CLIPIQA$\ \uparrow$ \\
        \hline
        $\checkmark$ &$\times$ &$\checkmark$ &20.86 &0.4564 &\underline{0.3926} &\underline{0.2362} &52.89 &0.5855 \\
        $\times$ &$\checkmark$ &$\checkmark$ &21.18 &0.4615 &0.4006 &0.2414 &\underline{53.36}  &\underline{0.5909} \\
        $\checkmark$ &$\times$ &$\times$ &\textbf{22.13} &\textbf{0.5092} &0.6280 &0.3739 &23.94 &0.1533 \\
        $\checkmark$ &$\checkmark$ &$\checkmark$ &\underline{21.33} &\underline{0.4691}  &\textbf{0.3852} &\textbf{0.2304} &\textbf{54.44} &\textbf{0.6014} \\
        \hline
	\end{tabularx}
\end{table*}

Next, we individually assess the effectiveness of the DiNO feature extractor and the semantic token aggregation module. Specifically, we experiment with replacing the DiNO-based feature extractor with the image encoder from CLIP to extract global features, and substituting the semantic token aggregation module with a simple MLP to perform feature dimension alignment. Table \ref{tab:ablation_global} presents the experimental results, demonstrating that DiNO outperforms CLIP in terms of feature extraction capability, achieving superior performance across all evaluation metrics. Compared to a simple MLP-based feature alignment, our semantic token aggregation module incurs almost no performance drop. However, it significantly reduces the number of tokens, thereby lowering the computational cost of the cross-attention layers within the UNet. The computational cost of the cross-attention mechanism in the UNet can be expressed as follows.
\begin{align}
\text{FLOPs}_{\text{attention}} = B \cdot H \cdot N_q \cdot N_k \cdot (4  d_h + 5)
\end{align}
Here, $B$, $H$, $N_{q}$ and $N_{k}$ denote the batch size, number of attention heads, number of query tokens, number of key/value tokens, and the dimension of each head, respectively. Our Semantic Token Aggregation module reduces $N_{k}$ from 900 to 96, resulting in an 89.3\% reduction in computational cost.

\begin{table}[htb]
    \caption{Ablation study of global branch. The bold and underlined numbers denote the best and second-best results.}
    \label{tab:ablation_global}
    \centering
    \footnotesize 
    \begin{tabularx}{\linewidth}{CCCCCCC}
        \hline
         &PSNR$\ \uparrow$ &SSIM$\ \uparrow$ &LPIPS$\ \downarrow$ &DISTS$\ \downarrow$ &MUSIQ$\ \uparrow$ &CLIPIQA$\ \uparrow$ \\
        \hline
        CLIP &20.97 &0.4557 &0.3917 &\underline{0.2320} &53.94 &0.5950 \\
        MLP &\textbf{21.37} &\underline{0.4645} &\underline{0.3885} &0.2371 &\underline{54.27} &\underline{0.5994} \\
        Ours &\underline{21.33} &\textbf{0.4691}  &\textbf{0.3852} &\textbf{0.2304} &\textbf{54.44} &\textbf{0.6014} \\
        \hline
    \end{tabularx}
\end{table}
\subsubsection{Effectiveness of Local Branch}
Our local branch is specifically designed to inject high-frequency local texture details from the reference image into the LR image. To validate its effectiveness, we replace the local texture encoder with a simple concatenation operation, where the LR and reference images are directly concatenated and passed through a series of convolutional and downsampling layers to extract features. As shown in the second row of Table \ref{tab:ablation_study}, this alternative yields inferior performance across both full-reference and no-reference metrics. This demonstrates that our proposed local texture encoder is more capable of extracting high-frequency textures that are relevant to the LR image, thereby enhancing the overall SR quality.

\begin{figure}[htb]
    \centering
    \includegraphics[width=\linewidth]{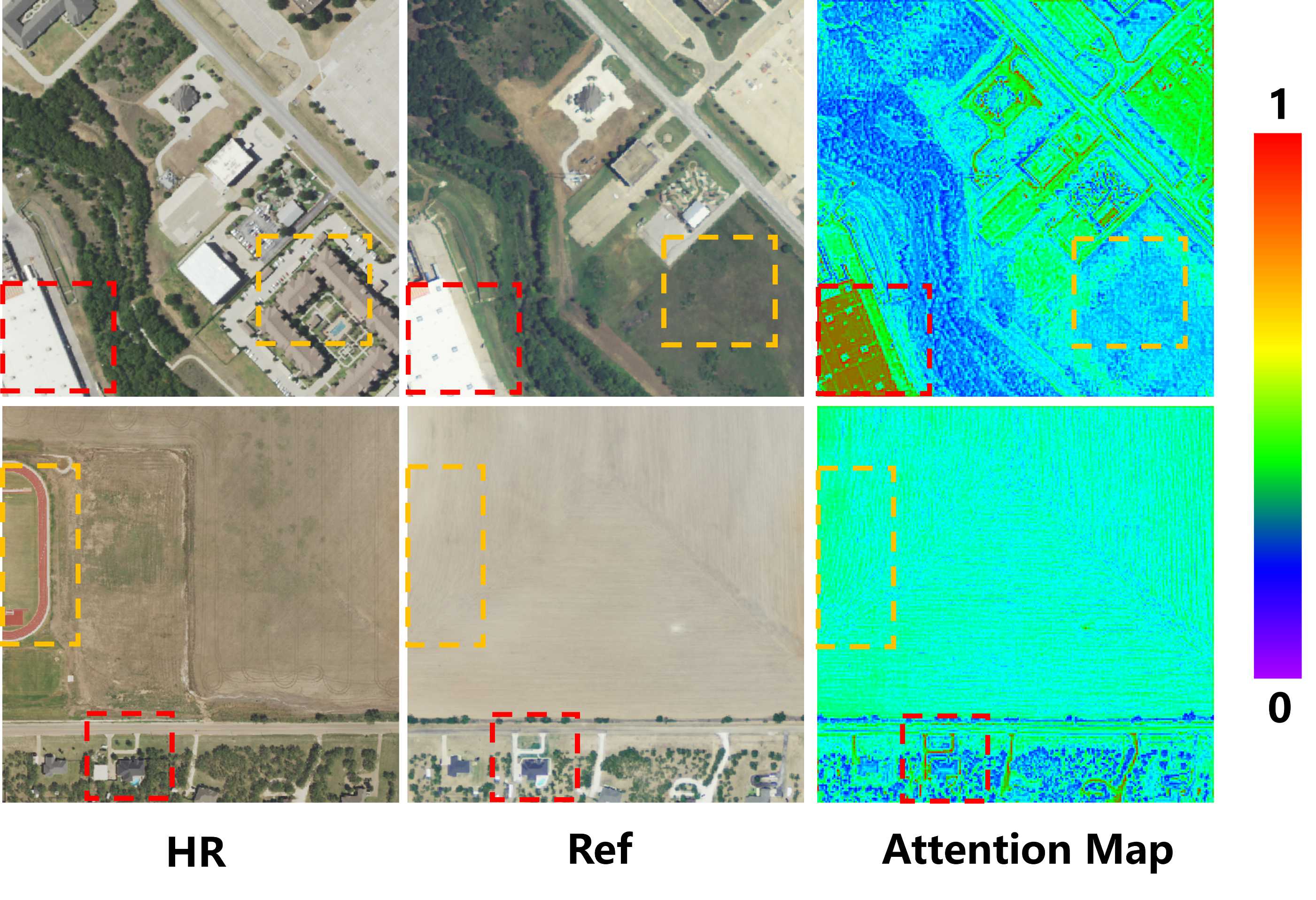}
    \caption{Visualization of attention maps in the Change Aware Attention (CAA) module.}
    \label{fig:local_attention}
\end{figure}

The core of the local branch is the Change Aware Attention (CAA) module, which adaptively integrates reference information based on local attention between the LR and reference images. As shown in Fig. \ref{fig:local_attention}, we visualize two sets of attention maps. It can be observed that in regions with noticeable land cover changes (highlighted by yellow boxes), the model tends to rely less on the reference image. In contrast, for unchanged yet complex regions such as buildings and roads (highlighted by red boxes), the model heavily utilizes information from the reference. This behavior is reasonable: in complex areas, the LR image alone provides limited information due to its low spatial resolution, prompting the network to prioritize the more informative reference image to enhance the reconstruction quality. For relatively homogeneous regions such as forests and farmland, the model combines information from both the LR and reference images for reconstruction. As a result, the attention values in these areas tend to be moderate, reflecting a balanced reliance on both sources.

\begin{figure}[htb]
    \centering
    \includegraphics[width=\linewidth]{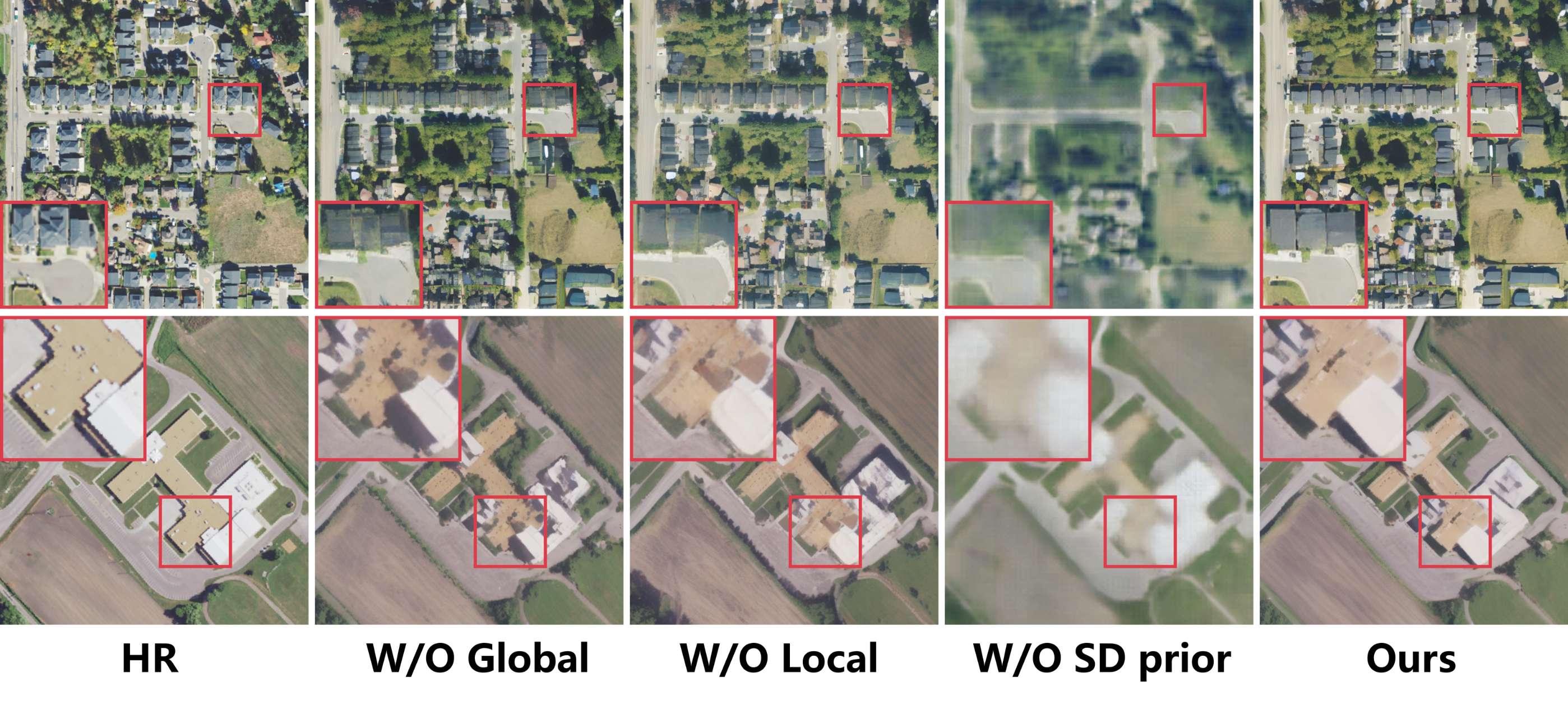}
    \caption{Visual comparison between our full model and its three ablated variants, each removing one of the following components: the global branch, the local branch, or the SD prior.}
    \label{fig:abla_visual}
\end{figure}
\subsubsection{Effectiveness of SD Prior}
Considering the substantial resolution gap between different satellite sensors in real-world scenarios, we incorporate the generative prior of a pretrained SD model to alleviate this ill-posed problem. Here, we experiment with removing the SD prior and relying solely on the local texture encoder as the backbone to perform the super-resolution task. As shown in the third row of Table 3, this configuration leads to a slight increase in PSNR but results in a significant degradation in perceptual quality metrics. In Fig. \ref{fig:abla_visual}, we present a visual comparison of different ablation settings. Removing either the global or local branch introduces color distortions and structural artifacts in the SR results. In contrast, without the SD prior, the model fails to recover high-frequency details. By comparison, our full model achieves superior fidelity in color reproduction, structural accuracy, and texture richness.

\subsubsection{Effectiveness of Augmentation Strategy}
\begin{figure}[htb]
    \centering
    \includegraphics[width=\linewidth]{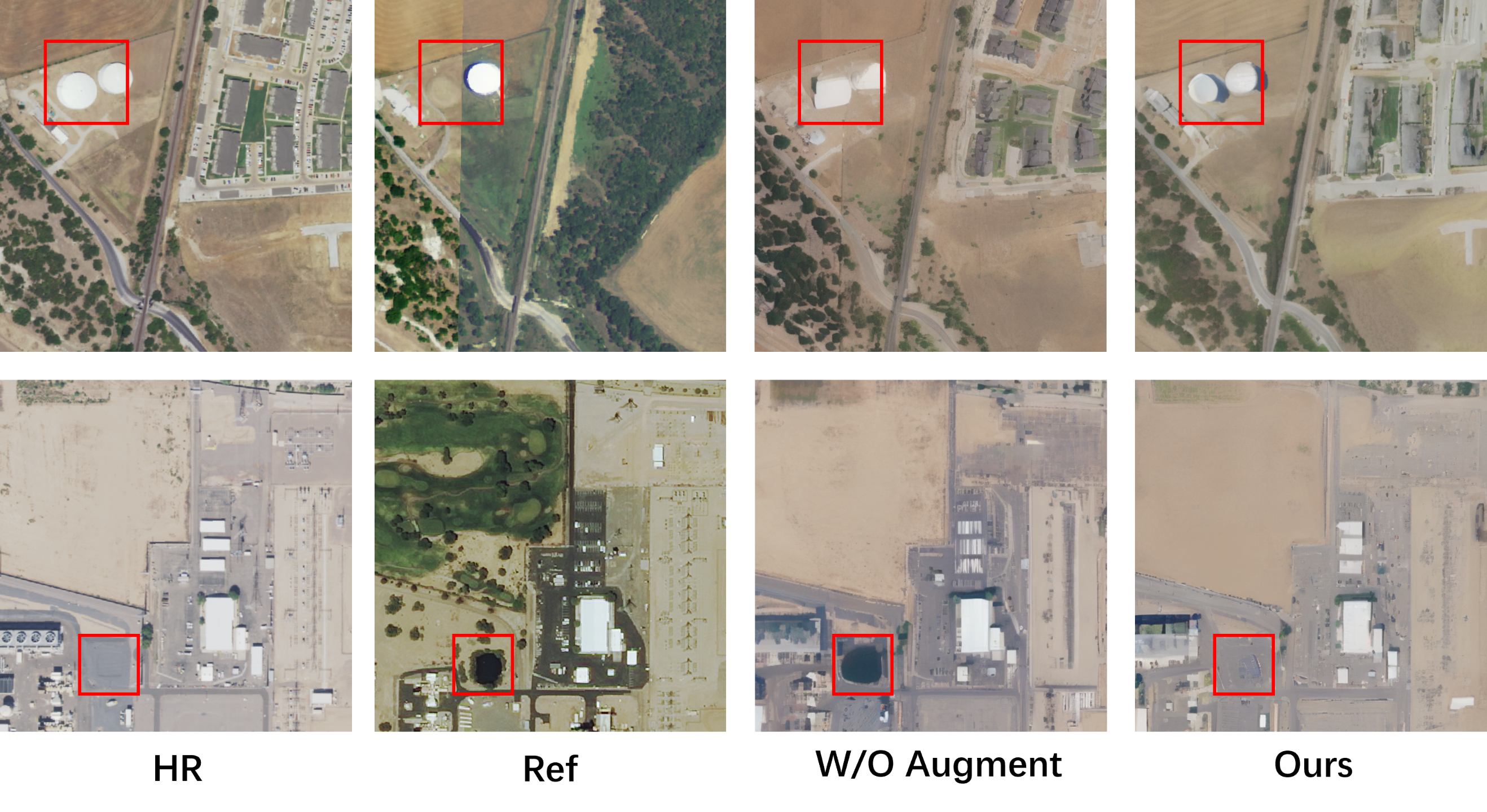}
    \caption{Visual comparison with and without the reference image augmentation strategy.}
    \label{fig:abla_aug}
\end{figure}

In this section, we validate the effectiveness of the employed reference image augmentation strategy. This strategy perturbs the style of reference images during training, thereby encouraging the model to focus more on structural content information. Fig. \ref{fig:abla_aug} presents reconstruction results without applying this strategy during training. It can be observed that, in this case, the model tends to overly depend on reference information, transferring non-existent objects into the reconstructed images. In addition, the quantitative results in Table \ref{tab:abla_aug} also demonstrate the performance gains achieved by this strategy. These findings collectively validate the effectiveness of the proposed strategy.

\begin{table}[htb]
	\caption{Quantitative comparison with and without the reference image augmentation strategy.}
	\label{tab:abla_aug}
	\centering
    \footnotesize
	\begin{tabularx}{\linewidth}{CCCCCCCC}
		\hline
         &PSNR$\ \uparrow$ &SSIM$\ \uparrow$ &LPIPS$\ \downarrow$ &DISTS$\ \downarrow$ &MUSIQ$\ \uparrow$ &CLIPIQA$\ \uparrow$ \\
        \hline
        W/O Augment &21.06 &0.4581 &0.3951 &0.2371 &54.14 &0.5978 \\
        Ours &\textbf{21.33} &\textbf{0.4691}  &\textbf{0.3852} &\textbf{0.2304} &\textbf{54.44} &\textbf{0.6014} \\
        \hline
	\end{tabularx}
\end{table}

\subsubsection{Better Start for Accelerated Inference}
\begin{figure}[htb]
    \centering
    \includegraphics[width=\linewidth]{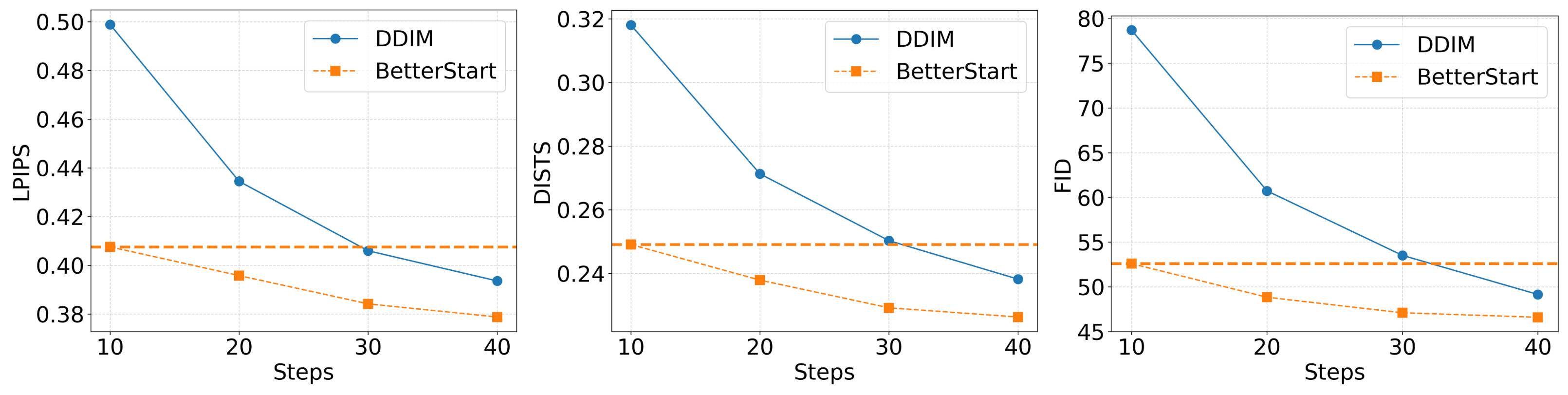}
    \caption{Comparison of Better Start strategy and DDIM sampling for inference acceleration. For these three metrics, lower values indicate better performance.}
    \label{fig:BetterStart}
\end{figure}

The inference process of diffusion models typically requires numerous iterative steps, resulting in substantial computational overhead and resource consumption \citep{song2021denoising}. To address this issue, we propose a Better Start strategy aimed at reducing the number of required inference steps and improving overall efficiency. Unlike conventional approaches such as DDIM \citep{song2021denoising}, which accelerate inference by simply increasing the sampling interval, the Better Start strategy reduces the sampling trajectory by modifying the starting point of the reverse denoising process. As illustrated in Fig. \ref{fig:BetterStart}, we compare the performance of Better Start and DDIM under the same number of sampling steps. The results demonstrate that our strategy consistently outperforms DDIM at the same number of sampling steps. Moreover, we observe that the Better Start strategy achieves performance comparable to DDIM with 30 steps using only 10 steps, highlighting the effectiveness of our approach in enhancing both efficiency and reconstruction quality.

\subsection{Discussion}
\subsubsection{High Spatiotemporal Resolution Image Synthesis}
\begin{figure}[htb]
    \centering
    \includegraphics[width=\linewidth]{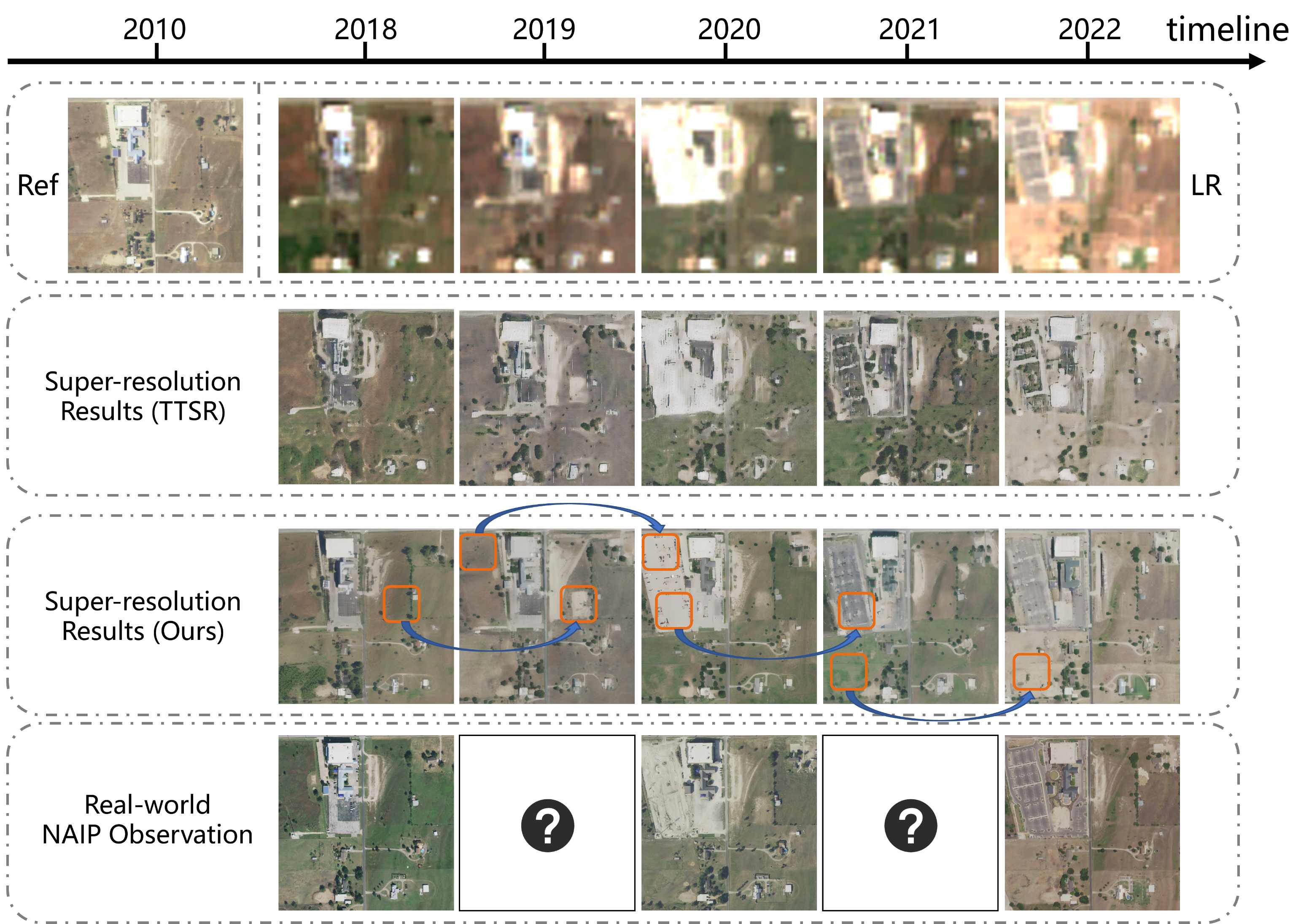}
    \caption{Illustration of our method for generating high spatiotemporal resolution imagery.}
    \label{fig:spa_temp_sys}
\end{figure}

For a given region, our model can generate HR observations at any time by leveraging LR inputs, conditioned on a historical HR observation. This enables more frequent and dynamic earth monitoring than what is achievable with existing satellite systems. Fig. \ref{fig:spa_temp_sys} illustrates this potential application. Our model can generate HR images for each year from 2018 to 2022, whereas NAIP provides HR observations for this area only in 2018, 2020, and 2022. The synthesized results accurately capture surface changes over time, such as the emergence of a plaza and the interchange between bare land and grassland. However, the reconstructed results from the competing method TTSR exhibit significant structural distortions and semantic inaccuracies. This further demonstrates the superior generalization ability and robustness of our model in real-world scenarios.

\subsubsection{Cross-Dataset Transfer Evaluation}
To evaluate the generalization capability of our model, we collected an additional cross-domain dataset for validation. In this dataset, both the HR images and the reference images are obtained from the National Ecological Observatory Network (NEON) with a spatial resolution of 1 m, while the LR images are acquired from Sentinel-2. The dataset was collected in the Edgewater region of the United States, which is not covered by our original dataset in either time or space. We directly evaluate the model trained on our proposed dataset on this cross-domain dataset. The results of our method and four representative RefSR baselines are presented in Table \ref{tab:cross_val}. As shown, our method consistently outperforms the comparison approaches, demonstrating superior performance even under cross-domain conditions. This indicates that our approach achieves stronger robustness and generalization in real-world scenarios.
\begin{table}[htb]
	\caption{Cross-dataset transfer evaluation results. The bold and underlined numbers denote the best and second-best results.}
	\label{tab:cross_val}
	\centering
	\begin{tabularx}{\linewidth}{CCCCC}
		\hline
         &LPIPS$\ \downarrow$ &DISTS$\ \downarrow$ &MUSIQ$\ \uparrow$ &CLIPIQA$\ \uparrow$ \\
        \hline
        TTSR &0.5319 &0.3420 &\underline{57.83} &0.6075 \\
        DATSR &0.6128 &0.3464 &37.35 &0.4617 \\
        RRSGAN &0.5501 &0.3372 &43.59 &0.5957 \\
        RGTGAN &\underline{0.4734} &\underline{0.2833} &55.13 &\underline{0.6389} \\
        Ours &\textbf{0.4203} &\textbf{0.2361} &\textbf{63.01} &\textbf{0.7424} \\
        \hline
	\end{tabularx}
\end{table}

\begin{figure*}[htb]
    \centering
    \includegraphics[width=\linewidth]{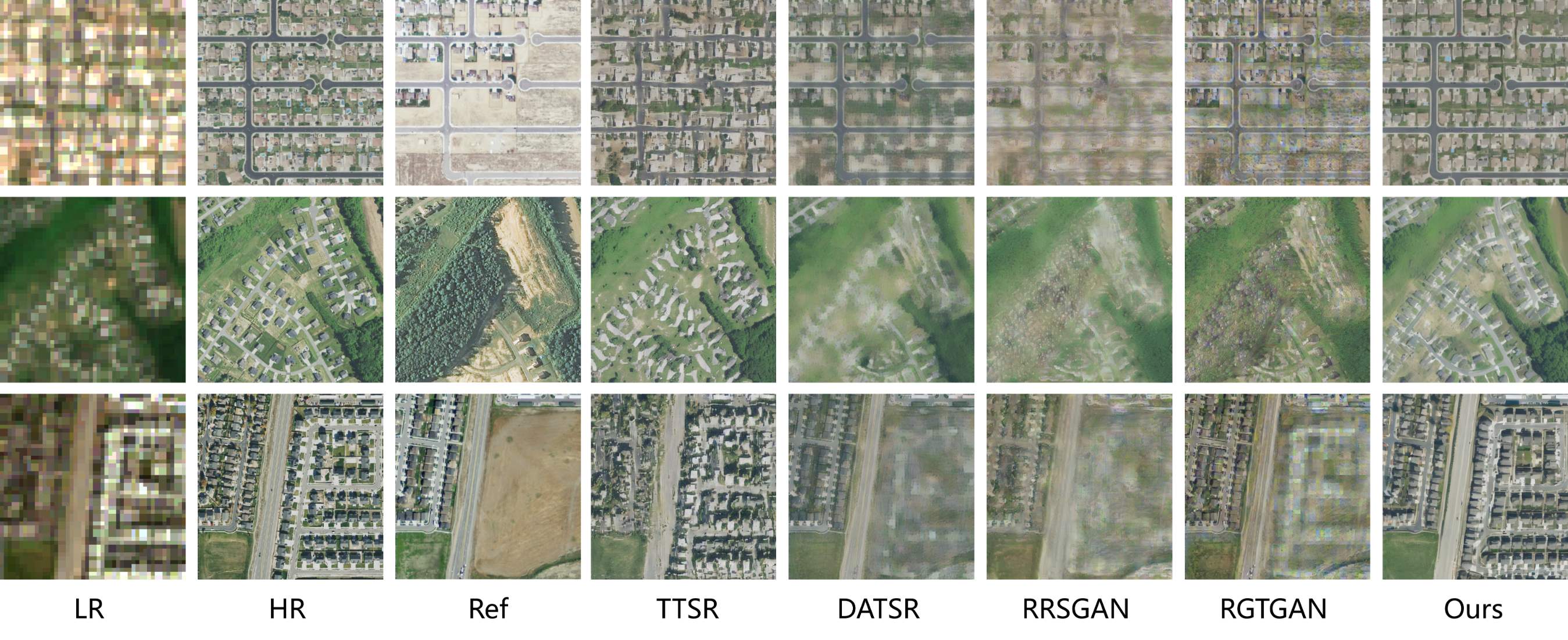}
    \caption{Visual comparison of reconstruction results between our method and other RefSR approaches under extreme land-cover changes.}
    \label{fig:extreme}
\end{figure*}

\subsubsection{Effect of Reference Similarity}
\begin{figure}[htb]
    \centering
    \includegraphics[width=\linewidth]{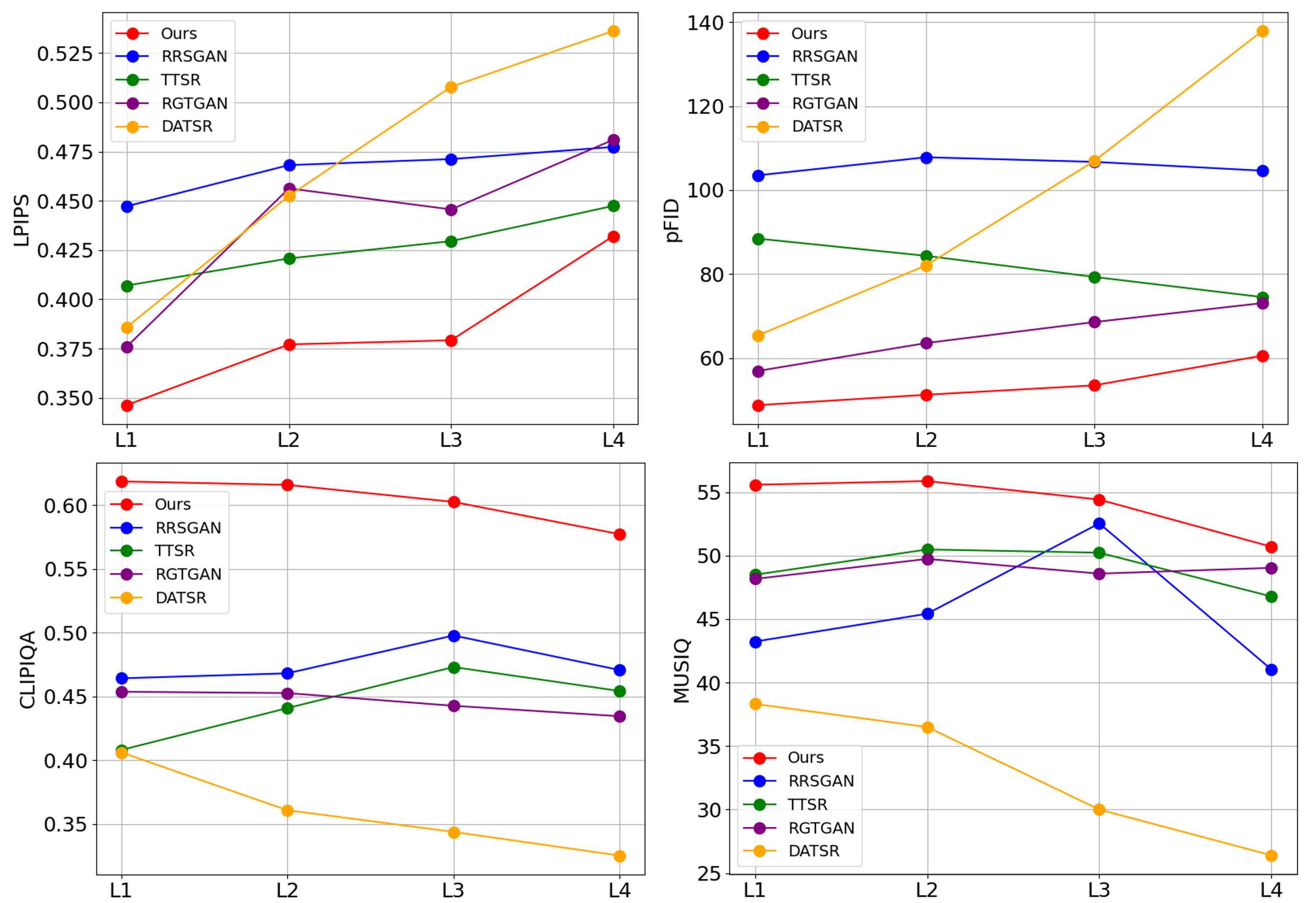}
    \caption{Performance comparison of different models across test subsets with varying levels of reference similarity. For CLIPIQA and MUSIQ, the higher, the better. While for LPIPS and pFID, the lower, the better.}
    \label{fig:LX_compare}
\end{figure}
In general, the similarity between the reference and the underlying ground-truth HR image has a potential impact on the performance of RefSR algorithms. To investigate the model's robustness under varying levels of reference similarity, we divide the entire test set into four subsets (L1–L4) based on the similarity degree between the reference and HR images, with L1 representing the highest similarity and L4 the lowest. We select four representative metrics to compare the performance of our method with all competing RefSR methods across different test subsets, as shown in Fig.~\ref{fig:LX_compare}. On one hand, as the similarity between the reference and HR images decreases, all models exhibit an overall performance drop. This is because less useful information can be extracted from the reference image, forcing the models to rely more heavily on the sparse content in the LR image. However, the quality of image reconstruction does not solely depend on the similarity between the reference and target images. It is also influenced by the complexity of the image content itself. As a result, some methods in the figure exhibit irregular trends across certain metrics. On the other hand, our method consistently outperforms all baselines across all subsets and metrics, demonstrating superior robustness and generalization under varying levels of reconstruction difficulty.

Fig. \ref{fig:extreme} presents the visual results of our method and all compared RefSR in scenarios with substantial land-cover changes. In the three representative examples, numerous new buildings appear in the HR images compared to the reference images. The dense and fine-grained structures of these buildings pose significant challenges for accurate reconstruction. Under such conditions, the baseline methods tend to produce either overly blurry or heavily distorted results. In contrast, our approach is able to effectively exploit the reference information and accurately recover the structural details of all objects.

\subsubsection{Local-Global Reference Strength Control}
\begin{figure}[htb]
    \centering
    \includegraphics[width=\linewidth]{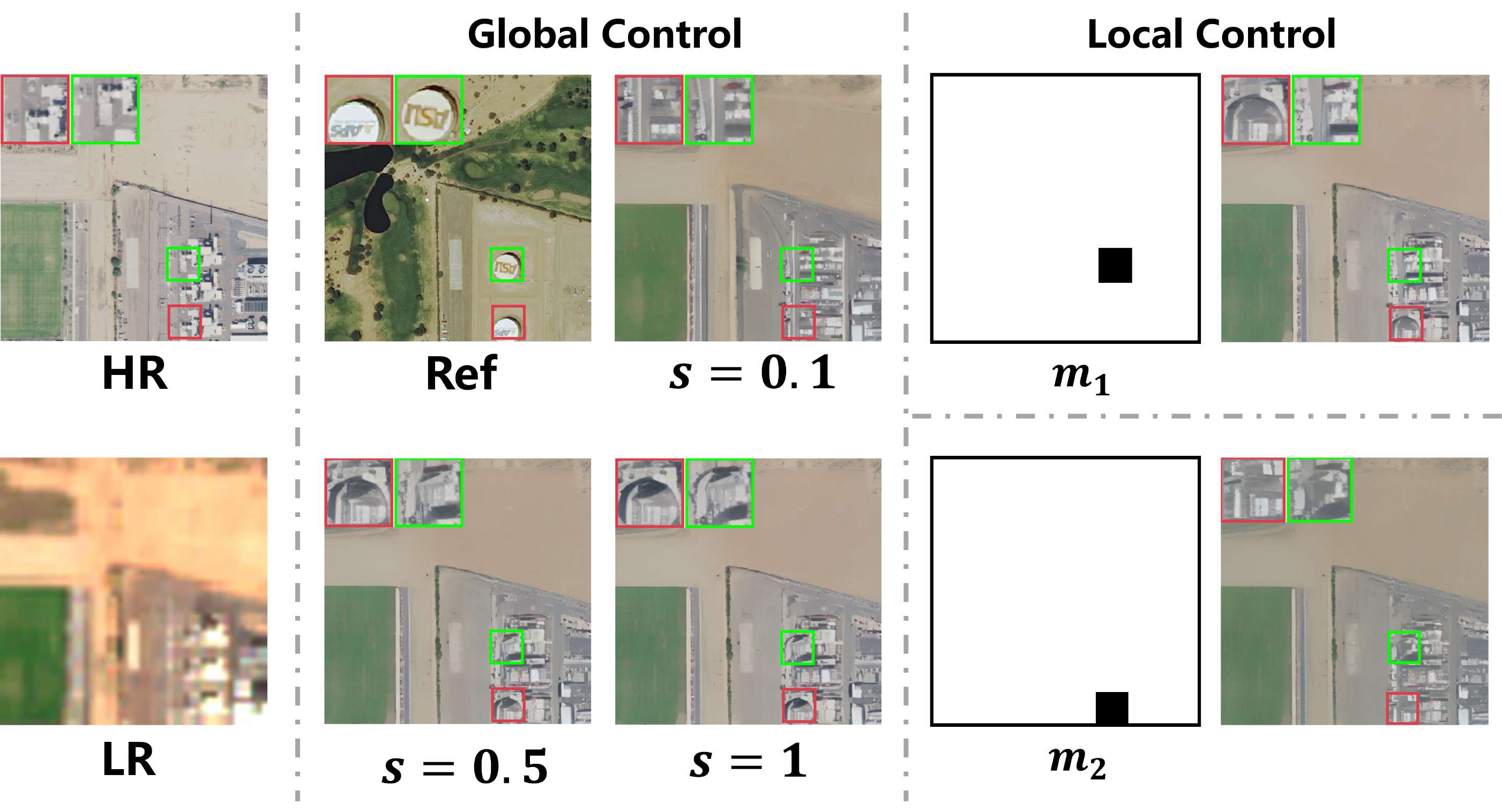}
    \caption{An illustrative example of global and local reference strength control.}
    \label{fig:control}
\end{figure}

Although our proposed CRefDiff demonstrates a strong ability to identify land cover changes between the LR and reference images, some failure cases involving incorrect reference usage still occur under certain challenging scenarios. To address this issue, we propose a Reference Strength Control strategy, which allows users to manually correct such cases. During inference, users can either supply a global scalar factor $s$ to adjust the overall reliance on reference information, or provide a spatial mask $m$ to modulate reference strength locally. As illustrated in Fig. \ref{fig:control}, a representative example is shown. In the highlighted region, the HR and reference images exhibit significant structural differences. However, the model still incorporates some misleading reference cues, resulting in erroneous structures in the reconstructed output (as indicated by ``$s=1$''). To alleviate this, the user can progressively reduce the global strength factor to suppress reference guidance. Alternatively, a binary mask specifying unreliable regions can be provided to locally weaken the influence of reference features in those areas.

\begin{table*}[htp]
    \centering
    \caption{Comparison of inference time and memory consumption.}
    \begin{tabularx}{\linewidth}{CCCCCCCCCC}
        \hline
        & TTST & ESRGAN & EDiffSR & DiffBIR & TTSR & DATSR & RRSGAN & RGTGAN & Ours \\ \hline
        Inference Time (ms) & 3586.4 & 42.197 & 24551 & 5780.5 & 96.772 & 1072.5 & 104.33 & 326.76 & 2840.6 \\ \hline
        Memory Consumption (M) & 85.39 & 72.71 & 146.6 & 6490 & 155.6 & 303.2 & 47.20 & 52.59 & 4256 \\ \hline
    \end{tabularx}
    \label{tab:effiency_comparison}
\end{table*}

\subsubsection{Model Efficiency}
To compare the runtime efficiency of different methods, we evaluate the inference time and GPU memory consumption for a 480$\times$480 image on an NVIDIA GeForce RTX 3090 GPU. As shown in Table \ref{tab:effiency_comparison}, our method exhibits a moderate inference time and relatively high memory usage. Specifically, TTST and DATSR require longer inference times due to the computational overhead of top-k attention mechanisms. EDiffSR and DiffBIR suffer from significantly reduced speed due to the iterative sampling nature of diffusion models. Regarding memory consumption, both our method and DiffBIR are built upon the pretrained Stable Diffusion model, resulting in a large number of parameters and increased memory usage. However, compared to the ControlNet-style architecture of DiffBIR , our adapter-based design is more lightweight, leading to lower memory consumption. The long inference time and massive memory consumption of SR methods based on naive Stable Diffusion models are common limitations. In the future, one-step distillation acceleration and model light-weighting techniques will be among our key directions for optimization.
\section{Conclusion}
In this paper, we cast remote sensing RefSR as a multi-sensor data fusion task and propose CRefDiff, a novel controllable reference-guided diffusion framework designed for real-world remote sensing image RefSR. To address the large resolution gap and complex land-cover variations commonly found in corss-sensor satellite imagery, CRefDiff integrates powerful generative priors from a pretrained generative model with a dual-branch fusion mechanism that captures both local textures and global contextual information from the reference image. Specifically, a Change-Aware Attention block is designed to selectively inject fine-grained local details, while a Semantic Token Aggregation module distills global features from DiNOv2 embeddings to provide contextual guidance. Moreover, we introduce two optional inference strategies---Better Start for faster denoising and reference strength control for adaptive reference information modulation---making the framework both efficient and interactive.

To support future research in this direction, we construct Real-RefRSSRD, the first real-world RefSR dataset featuring HR NAIP and LR Sentinel-2 pairs with diverse land-cover dynamics and temporal gaps. Extensive experiments show that CRefDiff achieves state-of-the-art performance on Real-RefRSSRD across multiple quantitative and perceptual metrics, and also brings tangible benefits to downstream tasks such as scene classification and semantic segmentation. These results underscore the potential of CRefDiff in advancing high spatiotemporal resolution image generation for remote sensing applications.
{
    \small
    \bibliographystyle{ieeenat_fullname}
    \bibliography{main}
}


\end{document}